\documentclass[10pt,twocolumn,letterpaper]{article}

\usepackage{cvpr}
\usepackage{times}
\usepackage{epsfig}
\usepackage{graphicx}
\usepackage{amsmath}
\usepackage{amssymb}
\usepackage{authblk}
\usepackage{multirow}
\usepackage{algorithm} 
\usepackage{algorithmic} 
\usepackage{xcolor}
\usepackage{paralist}
\usepackage{pifont}
\usepackage[misc]{ifsym}

\usepackage[breaklinks=true,bookmarks=false,colorlinks]{hyperref}

\cvprfinalcopy 

\def\cvprPaperID{751} 

\newcommand{\mypar}[1]{{\bf #1.}}
\ifcvprfinal\fi
\begin{document}

\title{Dynamic Multiscale Graph Neural Networks for \\ 3D Skeleton-Based Human Motion Prediction}

\author[1]{Maosen Li}
\author[2 \Letter]{Siheng Chen}
\author[1]{Yangheng Zhao}
\author[1 \Letter]{Ya Zhang}
\author[1]{Yanfeng Wang}
\author[3]{Qi Tian}

\affil[1]{{ Cooperative Medianet Innovation Center, Shanghai Jiao Tong University}}
\affil[2]{{ Mitsubishi Electric Research Laboratories}}
\affil[3]{{ Huawei Noah's Ark Lab}
\authorcr {\tt\small \{maosen\_li, zhaoyangheng-sjtu, ya\_zhang, wangyanfeng\} @sjtu.edu.cn, schen@merl.com, tian.qi1@huawei.com}}

\maketitle
\thispagestyle{empty}

\begin{abstract}
   We propose novel dynamic multiscale graph neural networks (DMGNN) to predict 3D skeleton-based human motions. The core idea of DMGNN is to use a multiscale graph to comprehensively model the internal relations of a human body for motion feature learning. This multiscale graph is adaptive during training and dynamic across network layers. Based on this graph, we propose a multiscale graph computational unit (MGCU) to extract features at individual scales and fuse features across scales. The entire model is action-category-agnostic and follows an encoder-decoder framework. The encoder consists of a sequence of MGCUs to learn motion features. The decoder uses a proposed graph-based gate recurrent unit to generate future poses. Extensive experiments show that the proposed DMGNN outperforms state-of-the-art methods in both short and long-term predictions on the datasets of Human 3.6M and CMU Mocap. We further investigate the learned multiscale graphs for the interpretability. The codes could be downloaded from \url{https://github.com/limaosen0/DMGNN}.
   \vspace{-12pt}
\end{abstract}

\section{Introduction}
3D skeleton-based human motion prediction forecasts future poses given the past motions based on the human-body-skeleton. The motion prediction helps machines understand human behaviors, attracting considerable attention~\cite{Fragkiadaki_2015_ICCV, Jain_2016_CVPR, Martinez_2017_CVPR, Butepage_2017_CVPR, Gui_2018_ECCV, Barsoum_2018_CVPR_Workshops}. The related techniques can be widely applied to many computer vision and robotics scenarios, such as human-computer interaction~\cite{7102751, pmlr-v28-koppula13, Huang_eccv_2014, gui-2018-110272}, autonomous driving~\cite{SihengChen}, and pedestrian tracking~\cite{Alahi_2016_CVPR, Gupta_2014_CVPR, Bhattacharyya_2018_CVPR}. 
\begin{figure}[t]
    \centering
    \includegraphics[width=0.9\columnwidth]{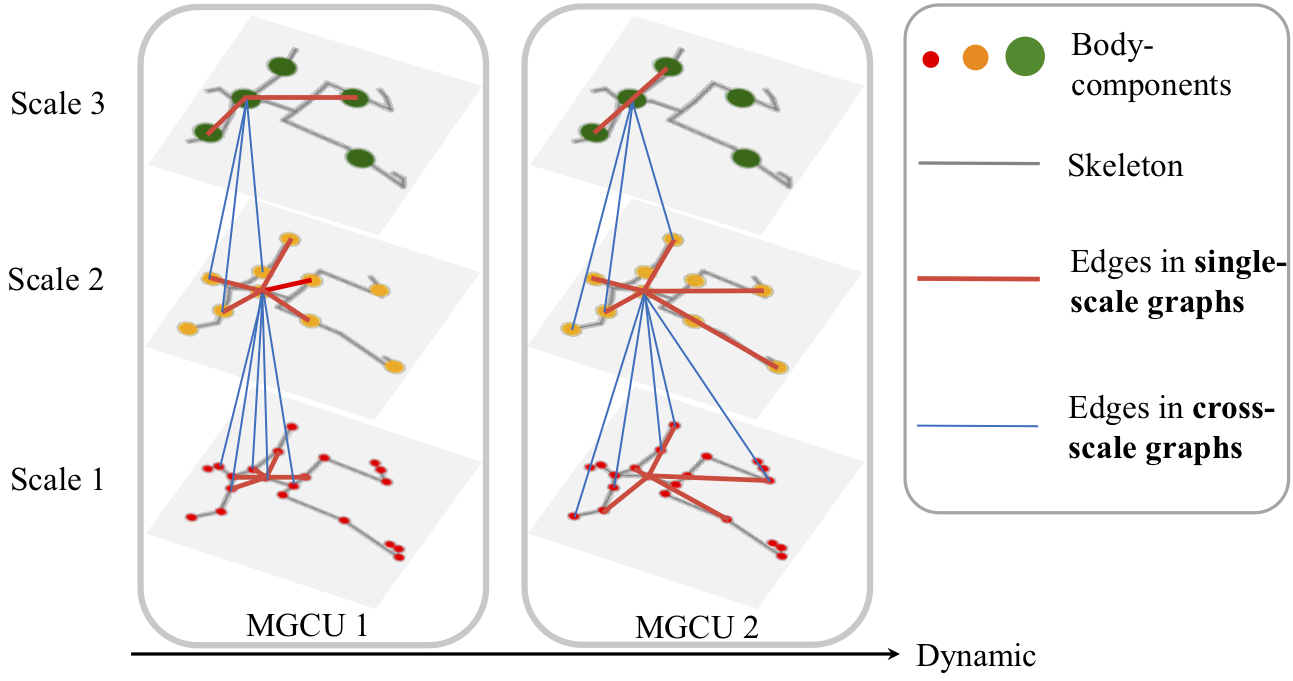}
    \small
    \caption{Two learned  multiscale graphs on `Posing'. We show strong relations associated with torsos in single scales and across scales. Two multiscale graphs are dynamic from one MGCUs to another, capturing local and distant relations, respectively.}
    \label{fig:jewel}
    \vspace{-5mm}
\end{figure}

Many methods, including the conventional state-based methods~\cite{Lehrmann_2014_CVPR, NIPS2005_2783, NIPS2006_3078, icml2009_129, NIPS2008_3567} and deep-network-based methods~\cite{Fragkiadaki_2015_ICCV, Martinez_2017_CVPR, GhoshSAH17, abs-1810-09676, Gui_2018_ECCV, AAAI_Guo, Gopalakrishnan_2019_CVPR, quater,WangSpatio}, have been proposed to achieve promising motion prediction. However, most methods did not explicitly exploit the relations or constraints between different body-components, which carry crucial information for motion prediction.  A recent work~\cite{Mao_2019_ICCV} built graphs across body-joints for pairwise relation modeling; however, such a graph was still insufficient to reflect a functional group of body-joints. Another work~\cite{WangSpatio} builds pre-defined sturctures to aggregate body-joint features to represent fixed body-parts, while the model only considers the body physical constraints without exploiting the movement coordination and relations. For example, the action of `Walking' tends to be understood based on the collaborative movements of abstract arms and legs, rather than the detailed locations of fingers and toes.
\begin{figure*}[t]
    \centering
    \includegraphics[width=1.9\columnwidth]{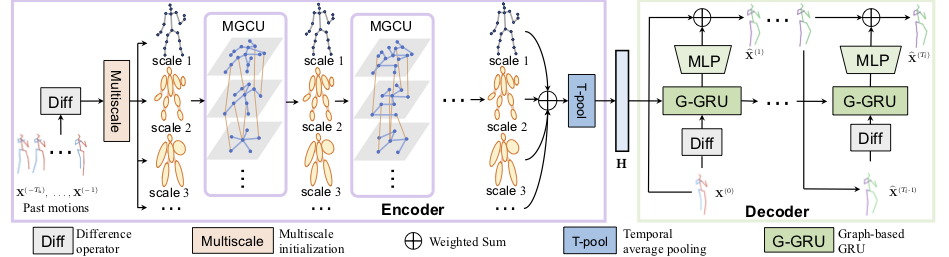}
    \caption{The architecture of DMGNN, which uses an encoder-decoder framework for motion prediction. In the encoder, cascaded multiscale graph computational blocks (MGCU) leverage dynamic muliscale graphs to extract spatio-temporal features. In the decoder, we propose a graph-based GRU (G-GRU) to predict poses.}
    \label{fig:pipeline}
    \vspace{-15pt}
\end{figure*}

To model more comprehensive relations, we propose a new representation for a human body: a~\emph{multiscale graph}, whose nodes are body-components at various scales and edges are pairwise relations between components. To model a body at multiple scales, a multiscale graph consists of two types of sub-graphs:~\emph{single-scale graphs}, connecting body-components at the same scales, and~\emph{cross-scale graphs}, connecting body-components across two scales; see Figure~\ref{fig:jewel}. The single-scale graphs together provide a pyramid representation of a body skeleton. Each cross-scale graph is a bipartite graph, bridging one single-scale graph to another. For example, an ``arm'' node in a coarse-scale graph could connect to  ``hand'' and ``elbow`` nodes in a fine-scale graph.  This multiscale graph is initialized by predefined physical connections and adaptively adjusted in training to be motion-sensitive. Overall, this multiscale representation provides a new potentiality to model body relations. 

Based on the multiscale graph, we propose a novel model, called~\emph{dynamic multiscale graph neural networks} (DMGNN), which is action-category-agnostic and follows from an encoder-decoder framework to learn motion representations for prediction. The encoder contains a cascade of~\emph{multiscale graph computational units} (MGCU), where each is associated with a multiscale graph. One MGCU includes two key components:~\emph{single-scale graph convolution block} (SS-GCB), leveraging single-scale graphs to exact features at individual scales, and~\emph{cross-scale fusion block} (CS-FB), inferring cross-scale graphs to convert features from one scale to another and enable fusion across scales. The multiscale graph has adaptive and trainable inbuilt topology; it is also dynamic because the topology is changing from one MGCU to another; see the learned dynamic multiscale graphs in Figure~\ref{fig:jewel}. 
Notably, cross-scale graphs in CS-FBs are constructed adaptively to input motions, and reflect discriminative motion patterns for category-agnostic  prediction. 

As for the decoder, we adopt a~\emph{graph-based gated recurrent unit} (G-GRU) to sequentially produce predictions given the last estimated poses. The G-GRU utilizes trainable graphs to further enhance state propagation. We also use residual connections to stabilize the prediction. To learn richer motion dynamics, we introduce difference operators to extract multiple orders of motion differences as the proxies of positions, velocities, and accelerations. The architecture of DMGNN is illustrated in Figure~\ref{fig:pipeline}.

To verify the superiority of our DMGNN, extensive experiments are conducted on two large-scale datasets: Human 3.6M~\cite{6682899} and CMU Mocap\footnote{http://mocap.cs.cmu.edu/}. The experimental results show that our model outperforms most state-of-the-art works for both short-term and long-term prediction in terms of both effectiveness and efficiency. The main contributions of this paper are as follow: 
\begin{itemize}
    \vspace{-2mm}
    \item We propose dynamic multiscale graph neural networks (DMGNN) to extract deep features at multiple scales and achieve effective motion prediction;
    \vspace{-2mm}
    \item We propose two key components: a multiscale graph computational unit, which leverages a multiscale graph to extract and fuse features across multiple scales, as well as a graph-based GRU to enhance state propagation for pose generation; and
    \vspace{-2mm}
    \item We conduct extensive experiments to show that the proposed DMGNN outperforms most state-of-the-art methods for short and long-term motion prediction on two large datasets. We further visualize the learned graphs for interpretability and reasoning.
\end{itemize}

\section{Related Work}
\textbf{Human motion prediction:} 
To forecast motions, some traditional methods, e.g., hidden Markov models~\cite{Lehrmann_2014_CVPR}, Gaussian-process~\cite{NIPS2005_2783} and random forests~\cite{Lehrmann_2014_CVPR}, were developed. Recently, deep networks are playing increasingly crucial roles: some recurrent-network-based models generated future poses step-by-step~\cite{Fragkiadaki_2015_ICCV, Jain_2016_CVPR, Martinez_2017_CVPR,  Walker_2017_ICCV, NIPS2016_6552, Gopalakrishnan_2019_CVPR, Liu_2019_CVPR,Gui_2018_ECCV, SymGNN}; some feed-forward networks~\cite{Li_2018_CVPR, Mao_2019_ICCV} tried to reduce error accumulation for stable prediction; imitation-learning algorithm was also proposed~\cite{Wang_2019_ICCV}. However, these methods rarely considered enough relations from various scales, which carry comprehensive information for human behaviors understanding. In this work, we build dynamic multiscale graphs to capture rich multiscale relations and extract flexible semantics for motion prediction. 

\textbf{Graph deep learning:} Graphs, expressing data associated with non-grid structures, preserve the dependencies among internal nodes~\cite{AAAI1817135, Verma_2018_CVPR, valsesia2018learning}. Many studies focused on graph representation learning and the relative applications~\cite{Li2018learning, NIPS2016_6081, kipf_iclr2017, NIPS2017_6703, AAAI1817135, Si_2018_ECCV}.  Based on fixed graph structures, previous works explored propagating node features according to either the graph spectral domain~\cite{NIPS2016_6081, kipf_iclr2017} or the graph vertex domain~\cite{NIPS2017_6703}. Several graph-based models have been employed for skeleton-based action recognition~\cite{AAAI1817135, Li_cvpr_2019,Shi_2019_CVPR}, motion prediction~\cite{Mao_2019_ICCV} and 3D pose estimation~\cite{Zhao_2019_CVPR}; Different from any previous works, our model considers multiscale graphs and corresponding operations.
\begin{figure}[t]
    \centering
    \includegraphics[width=0.85\columnwidth]{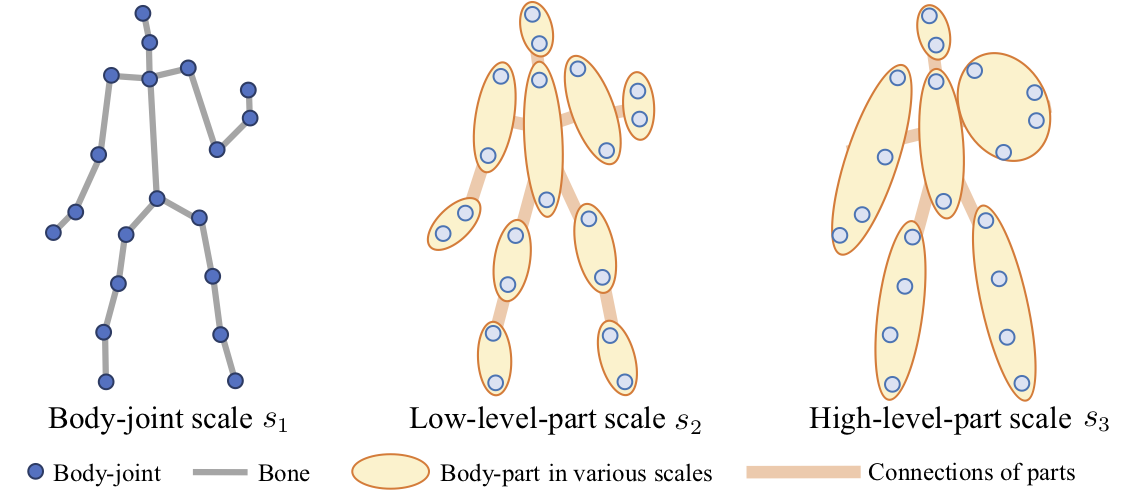}
    \small
    \caption{Three body scales on Human 3.6M. In $s_1$, we consider 20 joints with non-zero exponential maps~\cite{Huynh2009}; In $s_2$ and $s_3$, we consider 10 and 5 parts, respectively.}
    \label{fig:multiview_graphs}
    \vspace{-15pt}
\end{figure}

\section{Problem Formulation}
Suppose that the historical 3D skeleton-based poses are $\mathbb{X}_{-T_{\rm h}:0} = [\mathbf{X}^{(-T_{\rm h})},\dots,\mathbf{X}^{(0)}] {\in \mathbb{R}^{M \times (T_{\rm h}+1) \times D_{\bf x}}}$ and the future poses are $\mathbb{X}_{1:T_{\rm f}}=[\mathbf{X}^{(1)},\dots,\mathbf{X}^{(T_{\rm f})}] {\in \mathbb{R}^{M \times T_{\rm f} \times D_{\bf x}}}$, where $\mathbf{X}^{(t)}\in\mathbb{R}^{M\times D_{\bf x}}$ with $M$ joints and $D_{\bf x}=3$ feature-dimensions depicts the 3D pose at time $t$. The goal of motion prediction is to generate future poses given the past observed ones; mathematically, we need to propose a model $\mathcal{F}_{pred}(\cdot)$ to predict $\widehat{\mathbb{X}}_{1:T_{\rm f}}=\mathcal{F}_{pred}(\mathbb{X}_{-T_{\rm h}:0})$, where $\widehat{\mathbb{X}}_{1:T_{\rm f}}$ is the predicted motion close to the target $\mathbb{X}_{1:T_{\rm f}}$.

To exploit rich body relations, we represent a body as a multiscale graph across multiscale body-components. Theorically, we could use arbitrary number of scales. Based on human nature, we specifically adopt $3$ scales: the body-joint scale, the low-level-part scale, and the high-level-part scale. To initialize multiscale body graphs, we merge spatially nearby joints to coarser scales based on human prior; see Figure~\ref{fig:multiview_graphs}. With the multiscale graphs, we propose \emph{dynamic multiscale graph neural networks} (DMGNN) to predict future poses in an end-to-end fashion.

\section{Key Components}
\label{sec:key_components}
To construct our dynamic multiscale graph neural networks (DMGNN), we consider three basic components: a multiscale graph computational unit (MGCU), a graph-based GRU (G-GRU), and a difference operator.
\begin{figure}[t]
    \centering
    \includegraphics[width=0.85\columnwidth]{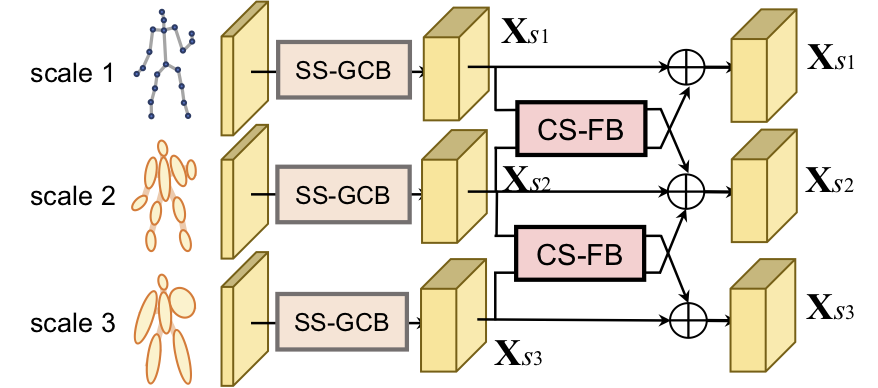}
    \small
    \caption{An MGCU uses single-scale graph convolution blocks (SS-CB) cross-scale fusion blocks (CS-FB).}
    \label{fig:MGCU}
    \vspace{-10pt}
\end{figure}

\subsection{Multiscale graph computational unit (MGCU)}
The functionality of a MGCU is to extract and fuse features at multiple scales based on a multiscale graph, which is trained adaptively and individually. One MGCU includes two types of building blocks:  single-scale graph convolution blocks, which leverage single-scale graphs to extract features at each scale, and cross-scale fusion blocks, which leverage cross-scale graphs to
convert features from one scale to another and enable effective fusion across scales; see Figure~\ref{fig:MGCU}. We now introduce each block in detail.

\mypar{Single-scale graph convolution block (SS-GCB)}
To extract spatio-temporal features at each scale, we propose a~\emph{single-scale graph convolution block} (SS-GCB). Let the trainable adjacency matrix of the single-scale graph at scale $s$ be $\mathbf{A}_{s}\in \mathbb{R}^{M_{s}\times M_{s}}$, where $M_{s}$ is the number of body-components. $\mathbf{A}_{s}$ is first initialized by a skeleton graph whose nodes are body-components and edges are physical connections, modeling a prior of the physical constraints; see Figure~\ref{fig:multiview_graphs}. During training, each element in ${{{\bf A}}_{s}}$ is adaptively tuned to capture flexible body relations. 

Based on the single-scale graph, SS-GCB effectively extracts deep features through two steps: 1) a graph convolution extracts spatial features of body-components; and 2) a temporal convolution extracts temporal features from motion sequences.  Let the input feature at scale $s$ be $\mathbf{X}_{s}\in\mathbb{R}^{M_{s}\times D_{\bf x}}$, the spatial graph convolution is formulated as
\vspace{-2mm}
\begin{equation}
\vspace{-2mm}
    \mathbf{X}_{s,{\rm sp}} = {\rm ReLU}(
                                           \mathbf{A}_{s}
                                           \mathbf{X}_{s}
                                           \mathbf{W}_{s} +
                                           \mathbf{X}_{s}
                                           \mathbf{U}_{s}) \in\mathbb{R}^{M_{s}\times D_{\bf x}'},
    \label{eq:spc_conv}
\end{equation}
where $\mathbf{W}_{s}, \mathbf{U}_{s}\in\mathbb{R}^{D_{\bf x} \times D_{\bf x}'}$ are trainable parameters. Through~\eqref{eq:spc_conv}, we extract the spatial features from correlated body-components. $\mathbf{A}_s$ in each SS-GCB is trained individually and stays fixed during test. To capture motions along time, we then develop a temporal convolution on the feature sequences.
The single-scale graphs in different SS-GCBs are dynamic, showing flexible relations. Note that features extracted at various scales have different dimensionalities and reflect 
information with different receptive fields.

\mypar{Cross-scale fusion block (CS-FB)}
To enable information diffusion across scales, we propose a~\emph{cross-scale fusion block} (CS-FB) which uses a cross-scale graph to convert features from one scale to another. A cross-scale graph is a bipartite graph that corresponds the nodes in one single-scale graph to the nodes in another single-scale graph. For example, the features of an ``arm'' node in the low-level-part scale $s_2$ can potentially guide the feature learning of a ``hand'' node in the body-joint scale $s_1$.  We aim to infer this cross-scale graph adaptively from data. Here we present  CS-FB from  $s_1$ to  $s_2$ as an example.

\begin{figure}[t]
    \centering
    \includegraphics[width=0.95\columnwidth]{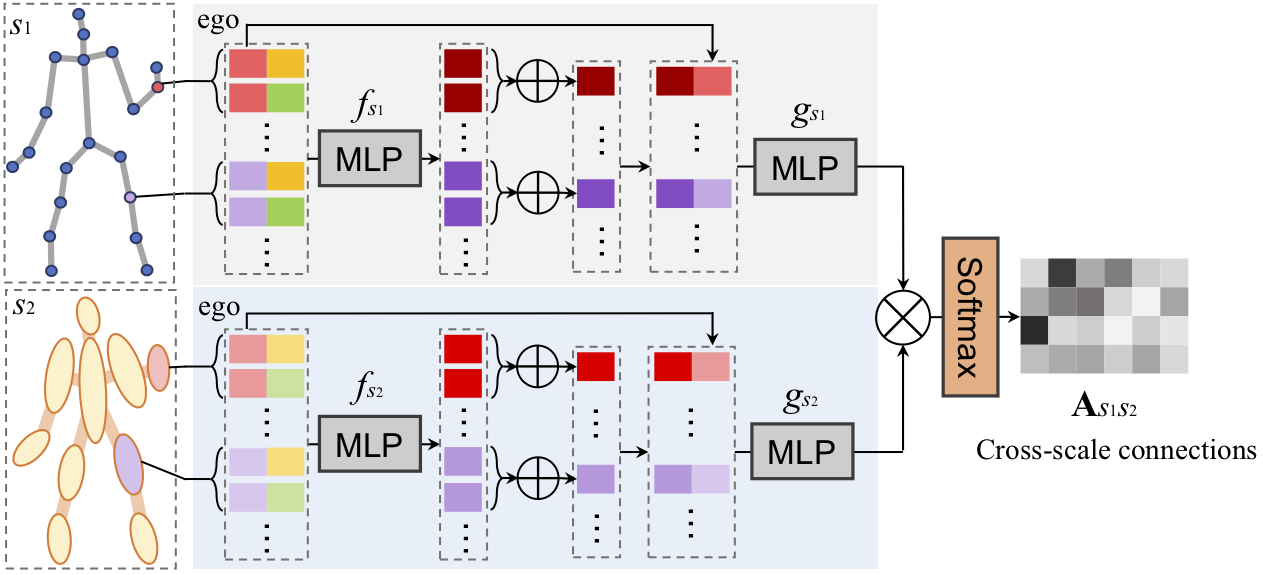}
    \small
    \caption{The inference of a cross-scale graph.}
    \label{fig:DC-FB}
    \vspace{-15pt}
\end{figure}

We first infer the cross-scale graph with adjacent matrix $\mathbf{A}_{s_1s_2} \in[0,1]^{M_{s_2}\times{M_{s_1}}}$ to model the cross-scale relations. Let the feature of the $i$th joint and the $k$th part along time be 
$(\mathbb{X}_{s_1} )_{:,i,:} \in \mathbb{R}^{T_{s_1} \times D_{\bf x}' }$ and 
$(\mathbb{X}_{s_2} )_{:,k,:} \in \mathbb{R}^{T_{s_2} \times D_{\bf x}' }$, we vectorize them as 
${\bf p}_{s_1,i}={\rm vec} ({\rm conv}_{s_1,\tau} ((\mathbb{X}_{s_1} )_{:,i,:};\mu))$ and 
${\bf p}_{s_2,k}={\rm vec} ({\rm conv}_{s_2,\tau} ((\mathbb{X}_{s_2} )_{:,k,:};\mu))$ to leverage temporal information, where $\tau$ and $\mu$ denote the temporal convolution kernel size and stride. We infer the edge weight between the $i$th joint and $k$th part $\left( \mathbf{A}_{s_1s_2} \right)_{k,i}$ through
\begin{subequations}
\vspace{-4mm}
\label{eq:nmp}
\begin{eqnarray}
\label{eq:j_nmp}
    \mathbf{r}_{s_1,i}  
    &=&   
    \sum_{j=1}^{M_{s_1}}
    f_{s_1}  
    \left(
         \left[
         {\bf p}_{s_1,i},
         {\bf p}_{s_1,j}-{\bf p}_{s_1,i}  \right]  
    \right) 
    \\
\label{eq:j_emb}
    {\bf h}_{s_1,i} 
    &=& 
    {\rm g}_{s_1} 
    \left(
    \left[{\bf p}_{s_1,i},
          {\bf r}_{s_1,i} 
    \right]
    \right)
    \\
\label{eq:p_nmp}
    \mathbf{r}_{s_2,k}  
    &=&
    \sum_{j=1}^{M_{s_2}}
    f_{s_2}  
    \left(
         \left[
         {\bf p}_{s_2,k},
         {\bf p}_{s_2,j}-{\bf p}_{s_2,k}  \right]  
    \right) 
    \\
\label{eq:p_emb}
    {\bf h}_{s_2,k} 
    &=& 
    {\rm g}_{s_2} 
    \left(
    \left[{\bf p}_{s_2,k},
          {\bf r}_{s_2,k} 
    \right]
    \right)
     \\
\label{eq:edge}
   \left( \mathbf{A}_{s_1s_2} \right)_{k,i} 
    &=&
    \mathrm{softmax}(
    \mathbf{h}_{{s_2}, k}^{\top}
    \mathbf{h}_{{s_1}, i}
    )\in [0,1],
\end{eqnarray}
\end{subequations}
where $f_{s_1} (\cdot)$, $g_{s_1} (\cdot)$, $f_{s_2}(\cdot)$ and $g_{s_2}(\cdot)$ denotes MLPs; $\mathrm{softmax}(\cdot)$ is a softmax operator along the raw of inner product matrix and $[\cdot,\cdot]$ is concatenation.~\eqref{eq:j_nmp} and~\eqref{eq:p_nmp} aggregate the relative features of all the components to the $i$th and the $k$th components in two scales, which are then updated by~\eqref{eq:j_emb} and~\eqref{eq:p_emb}; and~\eqref{eq:edge} obtains adjacent matrix through inner product and softmax, thus we model the normalized effects from a body in $s_1$ to each component in $s_2$. The intuition behind this design is to leverage the global relative information to augment body-component features, and we use the inner product of two augmented features to obtain the edge weight. Figure~\ref{fig:DC-FB} illustrates the inference of $\mathbf{A}_{s_1s_2}$. Notably, different from the fixed single-scale graphs during inference, the cross-scale graphs are efficiently inferred online and adaptive to motion features, which are flexible to capture distinct patterns for individual inputs.

We next fuse the joint features to the part-scale with $\mathbf{A}_{s_1s_2}$. Given the joint features at a certain time stamp $\mathbf{X}_{s_1}\in\mathbb{R}^{M_{s_1}\times D_{\bf x}'}$, the part-scale feature is updated as
\vspace{-2mm}
\begin{equation*}
\vspace{-2mm}
    {\bf X}_{s_2}
    \leftarrow  \mathbf{A}_{s_1s_2} 
     \mathbf{X}_{s_1} 
    \mathbf{W}_{{\rm F},s_1} 
    +\mathbf{X}_{s_2}\in\mathbb{R}^{M_{s_2}\times D_{\bf x}'},
    \label{eq:global_part}
\end{equation*}
where $\mathbf{W}_{{\rm F},s_1}\in\mathbb{R}^{D_{\bf x}'\times D_{\bf x}'}$ is trainable. Thus, each body-part in $s_2$ adaptively absorbs detailed information from the corresponding joints in $s_1$. The fused ${\bf X}_{s_2}$ is fed into the SS-CB of the next MGCU in $s_2$. In the other way around, we can define the fusion from $s_2$ to $s_1$ with similar operations.

\subsection{Graph-based GRU}
The functionality of a graph-based GRU (G-GRU) is to learn and update hidden states with the guide of a graph. The key is to use a trainable graph to regularize the states, which are used to generate future poses. Let  $\mathbf{A}_{\rm H} \in \mathbb{R}^{M \times M}$ be the adjacent matrix of the inbuilt graph, which is initialized with the skeleton-graph and trained to build adaptive edges, and $\mathbf{H}^{(0)} \in \mathbb{R}^{M \times D_{\bf h}}$ be the initial state of G-GRU. At time $t>0$, G-GRU takes two inputs: the initial state, $\mathbf{H}^{(t)}$, and the online 3D skeleton-based information, $\mathbf{I}^{(t)} \in \mathbb{R}^{M \times d}$. Then, G-GRU$(\mathbf{I}^{(t)}, \mathbf{H}^{(t)})$ works as
\vspace{-2mm}
\begin{equation*}
\vspace{-2mm}
    \begin{aligned}
        \mathbf{r}^{(t)} &= \sigma(r_{\rm in}(\mathbf{I}^{(t)}) + 
                            r_{\rm hid}({\mathbf{A}}_{\rm H} \mathbf{H}^{(t)} \mathbf{W}_{\rm H})), \\
        \mathbf{u}^{(t)} &= \sigma(u_{\rm in}(\mathbf{I}^{(t)}) + 
                            u_{\rm hid}({\mathbf{A}}_{\rm H} \mathbf{H}^{(t)} \mathbf{W}_{\rm H})), \\
        \mathbf{c}^{(t)} &= {\rm tanh}(c_{\rm in}(\mathbf{I}^{(t)}) + 
                            \mathbf{r}^{(t)} \odot c_{\rm hid}({\mathbf{A}}_{\rm H} \mathbf{H}^{(t)} \mathbf{W}_{\rm H})), \\
        \mathbf{H}^{(t+1)} &= \mathbf{u}^{(t)} \odot \mathbf{H}^{(t)} + 
                            (1-\mathbf{u}^{(t)}) \odot \mathbf{c}^{(t)}, \\
    \end{aligned}
\end{equation*}
where $r_{\rm in}(\cdot)$, $r_{\rm hid}(\cdot)$, $u_{\rm in}(\cdot)$, $u_{\rm hid}(\cdot)$, $c_{\rm in}(\cdot)$ and $c_{\rm hid}(\cdot)$ are trainable linear mappings; $\mathbf{W}_{\rm H}$ denotes the trainable weights.  For each G-GRU cell, it applies a graph convolution on the hidden states for information propagation and produces the state for next frame. 

\subsection{Difference operator} The motion states like velocity and acceleration carry important dynamics. To use them, we propose a difference operator to compute high-order differences of input sequences, guiding the model to learn richer dynamics. At time $t$, the $0$-order difference is $\Delta^{0}\mathbf{X}^{(t)}  = \mathbf{X}^{(t)}  \in \mathbb{R}^{M \times D_{\bf x}}$, and the $\beta$-order difference ($\beta > 0$) of the pose, $\Delta^{\beta}\mathbf{X}^{(t)}$, is 
$
    \Delta^{\beta}\mathbf{X}^{(t)} = \Delta^{\beta-1}\mathbf{X}^{(t)}-\Delta^{\beta-1}\mathbf{X}^{(t-1)}.
$
We use zero paddings after computing the differences to handle boundary conditions. Overall, the difference operator works as
\vspace{-3mm}
\begin{equation*}
\vspace{-1mm}
{\rm diff}_{\beta}(\mathbf{X}^{(t)})  =   \begin{bmatrix} \Delta^{0}\mathbf{X}^{(t)} & \cdots & \Delta^{\beta}\mathbf{X}^{(t)} \end{bmatrix}.
\end{equation*}
Here we consider $\beta=2$. The three elements reflects positions, velocities, and accelerations.

\section{DMGNN Framework}
Here we present the architecture of our DMGNN, which contains a multiscale graph-based encoder and a recurrent graph-based decoder for motion prediction.

\subsection{Encoder}
Capturing semantics from observed motions, the encoder aims to provide the decoder with motion states for prediction. In the encoder, for each motion sample, we first concatenate its $0,1,2$-order of differences as input. And we initialize $3$ body scales by averaging joint clusters in $s_1$ to spatially corresponding components in coarser scales. For example, we average two ``right hand'' joints in $s_1$ to the ``right arm'' part in $s_2$.  We then use a cascade of MGCUs to extract spatio-temporal features. Note that the multiscale graph associated with each MGCU is trained individually, thus the graph topology can be dynamically changing from one MGCU to another. To finally combine the three scales for comprehensive semantics, the output features are weighted summed. Since the numbers of body-components are different across scales, we broadcast the coarser components to match their spatially corresponding joints. Let the broadcast output features of the three scale be $\mathbb{H}_{s_1}, \mathbb{H}_{s_2}, \mathbb{H}_{s_3}\in\mathbb{R}^{T'\times M \times D_{\bf h}}$, the summed feature is
\vspace{-1mm}
\begin{equation}
\vspace{-2mm}
    \mathbb{H}=\mathbb{H}_{s_1}+
               \lambda(\mathbb{H}_{s_2}+\mathbb{H}_{s_3}),
    \label{eq:final_fusion}
\end{equation}
where $\lambda$ is a hyper-parameter to balance different scales. We next use a temporal average pooling to remove the time dimension of $\mathbb{H}$ and obtain $\mathbf{H} \in \mathbb{R}^{M \times D_{\bf h}}$, which aggregates historical information as the initial state of the decoder. 

\subsection{Decoder} 
The decoder aims to predict future poses sequentially. The core of the decoder is the proposed graph-based GRU (G-GRU), which further propagates motion states for sequence regression. We first use the difference operator to extract three orders of differences as motion priors, and then feed them into G-GRU to update the hidden state. We next 
generate future pose displacement with an output function. Finally, we add the displacements to the input pose to predict the next frame. At frame $t$, the decoder works as
\vspace{-2mm}
\begin{equation*}
\vspace{-2mm}
    \widehat{\mathbf{X}}^{(t+1)} = \widehat{\mathbf{X}}^{(t)} + f_{\rm pred}\left(\textrm{G-GRU}\left( {\rm diff}_{2}(\mathbf{\widehat{X}}^{(t)}), \mathbf{H}^{(t)}\right)\right),
\end{equation*}
where $f_{\rm pred}(\cdot)$ represents an output function, implemented by MLPs. The initial state $\mathbf{H}^{(0)} = \mathbf{H}$, which is the final output of encoder.  

\subsection{Loss function}
To train our DMGNN, we consider the $\ell_1$ loss. Let the $n$th sample of predictions be $(\widehat{\mathbb{X}}_{1:T_{\rm f}})_n \in \mathbb{R}^{T_{\rm f}\times M\times D_{\bf x}}$ and the corresponding ground truth be $(\mathbb{X}_{1:T_{\rm f}})_n$. For $N$ training samples, the loss function is
\vspace{-2mm}
\begin{equation*}
\vspace{-2mm}
\mathcal{L}_{\rm pred} =  \frac{1}{N} \sum_{n=1}^{N}
\left\|(\mathbb{X}_{1:T_{\rm f}})_n - (\widehat{\mathbb{X}}_{1:T_{\rm f}})_n \right\|_1,
\end{equation*}
where $||\cdot||_1$ denotes the  $\ell_1$ norm. $\ell_1$ loss gives sufficient gradients to joints with small losses to promote even more precise prediction; $\ell_1$ loss also gives stable gradients to joints with large losses, alleviating gradient explosion. In our experiments, $\ell_1$ loss leads to more precise predictions than $\ell_2$ loss. All the weights in the proposed DMGNN are trained end-to-end with the stochastic gradient descent~\cite{SGD}.

\begin{table*}[htb]
    \centering
    \caption{Mean angle errors (MAE) of different methods for short-term prediction on 4 representative actions of H3.6M. We also present different DMGNN variants, including using fixed graphs in SS-GCB (fixed $\mathbf{A}_{s}$), no graph in GRU (no G-GRU), and only one scale (single). The complete DMGNN outperform others methods at most time stamp.}
    \footnotesize
    \setlength{\tabcolsep}{2mm}{

        \begin{tabular}{c|cccc|cccc|cccc|cccc}
        \hline
        Motion & \multicolumn{4}{|c|}{Walking} & \multicolumn{4}{|c|}{Eating} & \multicolumn{4}{|c|}{Smoking} & \multicolumn{4}{|c}{Discussion}\\
        \hline
        milliseconds & 80&160&320&400 & 80&160&320&400 & 80&160&320&400 & 80&160&320&400 \\
        \hline
        ZeroV~\cite{Martinez_2017_CVPR} & 0.39 & 0.68 & 0.99 & 1.15 & 0.27 & 0.48 & 0.73 & 0.86 & 0.26 & 0.48 & 0.97 & 0.95 & 0.31 & 0.67 & 0.94 & 1.04 \\
        Res-sup.~\cite{Martinez_2017_CVPR} & 0.27 & 0.46 & 0.67 & 0.75 & 0.23 & 0.37 & 0.59 & 0.73 & 0.32 & 0.59 & 1.01 & 1.10 & 0.30 & 0.67 & 0.98 & 1.06 \\
        CSM~\cite{Li_2018_CVPR} & 0.33 & 0.54 & 0.68 & 0.73 & 0.22 & 0.36 & 0.58 & 0.71 & 0.26 & 0.49 & 0.96 & 0.92 & 0.32 & 0.67 & 0.94 & 1.01 \\
        TP-RNN~\cite{abs-1810-09676} & 0.25 & 0.41 & 0.58 & 0.65 & 0.20 & 0.33 & 0.53 & 0.67 & 0.26 & 0.47 & 0.88 & 0.90 & 0.30 & 0.66 & 0.96 & 1.04 \\
        AGED~\cite{Gui_2018_ECCV} & 0.21 & 0.35 & 0.55 & 0.64 & 0.18 & 0.28 & 0.50 & 0.63 & 0.27 & 0.43 & 0.81 & 0.83 & 0.26 & {0.56} & {0.77} & {0.84} \\
        Skel-TNet~\cite{AAAI_Guo} & 0.31 & 0.50 & 0.69 & 0.76 & 0.20 & 0.31 & 0.53 & 0.69 & 0.25 & 0.50 & 0.93 & 0.89 & 0.30 & 0.64 & 0.89 & 0.98 \\
        Imit-L~\cite{Wang_2019_ICCV} & 0.21 & 0.34 & 0.53 & 0.59 & 0.17 & 0.30 & 0.52 & 0.65 & 0.23 & 0.44 & 0.87 & 0.85 & 0.23 & 0.56 & 0.82 & 0.91 \\
        Traj-GCN~\cite{Mao_2019_ICCV} & {\bf 0.18} & {0.32} & {\bf 0.49} & {\bf 0.56} & {\bf 0.17} & 0.31 & 0.52 & 0.62 & 0.22 & 0.41 & 0.84 & 0.79 & {\bf 0.20} & {\bf 0.51} & {\bf 0.79} & {\bf 0.86}\\
        \hline
        DMGNN (fixed $\mathbf{A}_{s}$) & 0.20 & 0.35 & 0.54 & 0.63 & 0.20 & 0.34 & 0.53 & 0.66 & 0.23 & 0.41 & 0.86 & 0.83 & 0.26 & 0.65 & 0.92 & 1.02\\
        DMGNN (no G-GRU) & 0.22 & 0.33 & 0.53 & 0.61 & 0.19 & 0.32 & 0.53 & 0.66 & 0.23 & 0.42 & 0.87 & 0.82 & 0.27 & 0.65 & 0.90 & 0.98\\
        DMGNN ($S=1$) & 0.20 & 0.33 & 0.54 & 0.60 & 0.18 & 0.31 & 0.52 & 0.62 & 0.22 & 0.41 & 0.83 & 0.80 & 0.25 & 0.64 & 0.95 & 1.00\\
        DMGNN & {\bf 0.18} & {\bf 0.31} & {\bf 0.49} & {0.58} & {\bf 0.17} & {\bf 0.30} & {\bf 0.49} & {\bf 0.59} & {\bf 0.21} & {\bf 0.39} & {\bf 0.81} & {\bf 0.77} & {0.26} & {0.65} & {0.92} & {0.99}\\
        \hline
        \end{tabular}}
    \label{tab:pred_h36m_4}
\end{table*}

\begin{table*}[htb]
    \centering
    \caption{MAEs of different methods for short-term motion prediction on other $11$ actions of H3.6M.}
    
    \footnotesize
    \setlength{\tabcolsep}{0.76mm}{

        \begin{tabular}{c|cccc|cccc|cccc|cccc|cccc|cccc}
        \hline
        Motion & \multicolumn{4}{|c|}{Directions} & \multicolumn{4}{|c|}{Greeting} & \multicolumn{4}{|c|}{Phoning} 
               & \multicolumn{4}{|c|}{Posing} & \multicolumn{4}{|c|}{Purchases} & \multicolumn{4}{|c}{Sitting}\\
        \hline
        millisecond & 80&160&320&400 & 80&160&320&400 & 80&160&320&400 & 80&160&320&400 & 80&160&320&400 & 80&160&320&400 \\
        \hline
        Res-sup~\cite{Martinez_2017_CVPR} & 0.41 & 0.64 & 0.80 & 0.92 & 0.57 & 0.83 & 1.45 & 1.60 & 0.59 & 1.06 & 1.45 & 1.60 & 0.45 & 0.85 & 1.34 & 1.56 & 0.58 & 0.79 & 1.08 & 1.15 & 0.41 & 0.68 & 1.12 & 1.33 \\
        CSM~\cite{Li_2018_CVPR} & 0.39 & 0.60 & 0.80 & 0.91 & 0.51 & 0.82 & 1.21 & 1.38 & 0.59 & 1.13 & 1.51 & 1.65 & 0.29 & 0.60 & 1.12 & 1.37 & 0.63 & 0.91 & 1.19 & 1.29 & 0.39 & 0.61 & 1.02 & 1.18 \\
        Traj-GCN~\cite{Mao_2019_ICCV} & {0.26} & {0.45} & 0.70 & 0.79 & {\bf 0.35} & {\bf 0.61} & {0.96} & {1.13} & {\bf 0.53} & {1.02} & 1.32 & 1.45 & 0.23 & 0.54 & 1.26 & 1.38 & {0.42} & 0.66 & 1.04 & 1.12 & 0.29 & 0.45 & 0.82 & {\bf 0.97} \\
        \hline
        DMGNN & {\bf 0.25} & {\bf 0.44} & {\bf 0.65} & {\bf 0.71} & {0.36} & {\bf 0.61} & {\bf 0.94} & {\bf 1.12} & {\bf 0.52} & {\bf 0.97} & {\bf 1.29} & {\bf 1.43} & {\bf 0.20} & {\bf 0.46} & {\bf 1.06} & {\bf 1.34} & {\bf 0.41} & {\bf 0.61} & {\bf 1.05} & {1.14} & {\bf 0.26} & {\bf 0.42} & {\bf 0.76} & {\bf 0.97} \\
        \hline
        \hline
        
        Motion & \multicolumn{4}{|c|}{Sitting Down} & \multicolumn{4}{|c|}{Taking Photo} & \multicolumn{4}{|c|}{Waiting} 
               & \multicolumn{4}{|c|}{Walking Dog} & \multicolumn{4}{|c|}{Walking Together} & \multicolumn{4}{|c}{Average}\\
        \hline
        millisecond & 80&160&320&400 & 80&160&320&400 & 80&160&320&400 & 80&160&320&400 & 80&160&320&400 & 80&160&320&400 \\
        \hline
        Res-sup.~\cite{Martinez_2017_CVPR} & 0.47 & 0.88 & 1.37 & 1.54 & 0.28 & 0.57 & 0.90 & 1.02 & 0.32 & 0.63 & 1.07 & 1.26 & 0.52 & 0.89 & 1.25 & 1.40 & 0.27 & 0.53 & 0.74 & 0.79 & 0.40 & 0.69 & 1.04 & 1.18\\
        CSM~\cite{Li_2018_CVPR} & 0.41 & 0.78 & 1.16 & 1.31 & 0.23 & 0.49 & 0.88 & 1.06 & 0.30 & 0.62 & 1.09 & 1.30 & 0.59 & 1.00 & 1.32 & 1.44 & 0.27 & 0.52 & 0.71 & 0.74 & 0.38 & 0.68 & 1.01 & 1.13 \\
        Traj-GCN~\cite{Mao_2019_ICCV} & {\bf 0.30} & {\bf 0.63} & {\bf 0.89} & {\bf 1.01} & {\bf 0.15} & {0.36} & {0.59} & {0.72} & 0.23 & 0.50 & 0.92 & 1.15 & 0.46 & 0.80 & {\bf 1.12} & {\bf 1.30} & {\bf 0.15} & 0.35 & 0.52 & {\bf 0.57} & {\bf 0.27} & 0.53 & 0.85 & 0.96 \\
        \hline
        DMGNN & {0.32} & {0.65} & {0.93} & {1.05} & {\bf 0.15} & {\bf 0.34} & {\bf 0.58} & {\bf 0.71} & {\bf 0.22} & {\bf 0.49} & {\bf 0.88} & {\bf 1.10} & {\bf 0.42} & {\bf 0.72} & {1.16} & {1.34} & {\bf 0.15} & {\bf 0.33} & {\bf 0.50} & {\bf 0.57} & {\bf 0.27} & {\bf 0.52} & {\bf 0.83} & {\bf 0.95} \\
        \hline
        \end{tabular}}
        \vspace{-10pt}
    \label{tab:pred_h36m_11}
    \end{table*}
\begin{table}[t]
    \centering
    \caption{MAEs of different methods for long-term prediction on the $4$ representative actions of H3.6M dataset.}
    \footnotesize
    \setlength{\tabcolsep}{0.73mm}{
    \begin{tabular}{c|cc|cc|cc|cc|cc}
    
        \hline
        Motion & 
        \multicolumn{2}{|c|}{Walking}& \multicolumn{2}{|c}{Eating}&
        \multicolumn{2}{|c|}{Smoking}& \multicolumn{2}{|c}{Discussion} & \multicolumn{2}{|c}{Average} \\ \hline
         
        milliseconds & 560 & 1k & 560 & 1k & 560 & 1k & 560 & 1k & 560 & 1k\\ \hline
        ZeroV~\cite{Martinez_2017_CVPR} & 1.35 & 1.32 
                                        & 1.04 & 1.38 
                                        & 1.02 & 1.69 
                                        & 1.41 & 1.96 
                                        & 1.21 & 1.59\\
        Res-sup.~\cite{Martinez_2017_CVPR} & 0.93 & 1.03 
                                           & 0.95 & 1.08 
                                           & 1.25 & 1.50 
                                           & 1.43 & 1.69
                                           & 1.14 & 1.33\\
        CSM~\cite{Li_2018_CVPR} & 0.98 & 0.92 
                                & 1.01 & 1.24 
                                & 0.97 & 1.62 
                                & 1.56 & 1.86
                                & 1.13 & 1.41\\ 
        AGED~\cite{Gui_2018_ECCV} & 0.78 & 0.91
                                  & 0.86 & {\bf 0.93}
                                  & 1.06 & {\bf 1.21}
                                  & {\bf 1.25} & {\bf 1.30} 
                                  & 0.99 & {\bf1.09} \\
        Skel-TNet~\cite{AAAI_Guo} & 0.94 & 0.92 
                                  & 0.97 & 1.23 
                                  & 0.99 & 1.59 
                                  & 1.51 & 1.82
                                  & 1.10 & 1.39 \\
        Imit-L~\cite{Wang_2019_ICCV} & 0.67 & 0.69 
                                     & 0.79 & 1.13 
                                     & 0.95 & 1.63 
                                     & 1.34 & 1.81
                                     & 0.94 & 1.32\\
        Traj-GCN~\cite{Mao_2019_ICCV} & {\bf 0.65} & {\bf 0.67} 
                                      & {0.76} & 1.12 
                                      & {0.87} & 1.57 
                                      & 1.33 & 1.70
                                      & {0.90} & 1.27\\ \hline
        DMGNN & 0.66 & 0.75 
              & {\bf 0.74} & {1.14} 
              & {\bf 0.83} & {1.52} 
              & {1.33} & {1.45}
              & {\bf 0.89} & {1.21}\\ \hline
    \end{tabular}}
    \label{tab:longterm}
    \vspace{-15pt}
\end{table}
\begin{table*}[t]
    \centering
    \caption{Comparisons of MAEs between our model and the state-of-the-art methods on the 8 actions of CMU Mocap dataset. We evaluate the model and present the MAEs at both short and long-term prediction time stamps. }
    \footnotesize
    \setlength{\tabcolsep}{1.32mm}{

        \begin{tabular}{c|cccc|c|cccc|c|cccc|c|cccc|c}
        \hline
        Motion & \multicolumn{5}{|c|}{Basketball} & \multicolumn{5}{|c|}{Basketball Signal} & \multicolumn{5}{|c|}{Directing Traffic} &  \multicolumn{5}{|c}{Jumping} \\ \hline
        milliseconds & 80 & 160 & 320 & 400 & 1000 & 80 & 160 & 320 & 400 & 1000 & 80 & 160 & 320 & 400 & 1000 & 80 & 160 & 320 & 400 & 1000\\ \hline
        Res-sup.~\cite{Martinez_2017_CVPR} & 0.49 & 0.77 & 1.26 & 1.45 & 1.77 & 0.42 & 0.76 & 1.33 & 1.54 & 2.17 & 0.31 & 0.58 & 0.94 & 1.10 & 2.06 & 0.57 & 0.86 & 1.76 & 2.03 & 2.42\\
        CSM~\cite{Li_2018_CVPR} & 0.36 & 0.62 & 1.07 & 1.17 & 1.95 & 0.33 & 0.62 & 1.05 & 1.23 & 1.98 & 0.26 & 0.58 & 0.91 & 1.04 & 2.08 & 0.38 & 0.60 & {1.36} & {1.58} & 2.05 \\
        Traj-GCN~\cite{Mao_2019_ICCV} & 0.33 & 0.52 & {\bf 0.89} & {\bf 1.06} & 1.71 & 0.11 & 0.20 & 0.41  & 0.53 & {\bf 1.00} & {\bf 0.15} & {0.32} & {\bf 0.52} & {\bf 0.60} & 2.00 & {\bf 0.31} & {\bf 0.49} & {\bf 1.23} & {\bf 1.39} & {1.80} \\
        \hline
        DMGNN & {\bf 0.30} & {\bf 0.46} & {\bf 0.89} & {1.11} & {\bf 1.66} & {\bf 0.10} & {\bf 0.17} & {\bf 0.31} & {\bf 0.41} & 1.26 & {\bf 0.15} & {\bf 0.30} & {0.57} & {0.72} & {\bf 1.98} & {0.37} & {0.65} & 1.49 & 1.71 & {\bf 1.79} \\ 
        \hline
        \hline
        Motion & \multicolumn{5}{|c|}{Running} & \multicolumn{5}{|c|}{Soccer} & \multicolumn{5}{|c|}{Walking} & \multicolumn{5}{|c}{Washing Window} \\ \hline
        milliseconds & 80 & 160 & 320 & 400 & 1000 & 80 & 160 & 320 & 400 & 1000 & 80 & 160 & 320 & 400 & 1000 & 80 & 160 & 320 & 400 & 1000\\ \hline
        Res-sup.~\cite{Martinez_2017_CVPR} & 0.32 & 0.48 & 0.65 & 0.74 & 1.00 & 0.29 & 0.50 & 0.87 & 0.98 & 1.73 & 0.35 & 0.45 & 0.59 & 0.64 & 0.88 & 0.31 & 0.47 & 0.74 & 0.93 & 1.37 \\
        CSM~\cite{Li_2018_CVPR} & 0.28 & 0.43 & {0.54} & 0.57 & {0.69} & 0.28 & 0.48 & 0.79 & 0.90 & 1.58 & 0.35 & 0.44 & 0.46 & 0.51 & 0.77 & 0.30 & 0.47 & 0.79 & 1.00 & 1.39 \\
        Traj-GCN~\cite{Mao_2019_ICCV} & 0.33 & 0.55 & 0.73 & 0.74 & 0.95 & {\bf 0.18} & {\bf 0.29} & {\bf 0.61}  & {\bf 0.71} & {\bf 1.40} & 0.33 & 0.45 & 0.49 & 0.53 & 0.61 & 0.22 & 0.33 & {\bf 0.57} & {\bf 0.75} & 1.20 \\
        \hline
        DMGNN & {\bf 0.19} & {\bf 0.31} & {\bf 0.47} & {\bf 0.49} & {\bf 0.64} & {0.22} & {0.32} & {0.79} & {0.91} & {1.54} & {\bf 0.30} & {\bf 0.34} & {\bf 0.38} & {\bf 0.43} & {\bf 0.60} & {\bf 0.20} & {\bf 0.27} & {0.62} & {0.81} & {\bf 1.09}\\ 
        \hline
    \end{tabular}}
    \label{tab:pred_cmu}
    \vspace{-10pt}
\end{table*}

\section{Experiments}
\subsection{Datasets and experimental setup}
{\bf Human 3.6m} (H3.6M).
H3.6M dataset~\cite{6682899} has $7$ subjects performing $15$ different classes of actions. There are $32$ joints in each subject, and we transform the joint positions into the exponential maps and only use the joints with non-zero values ($20$ joints remain). Along the time axis, we downsample all sequences by two. Following previous paradigms~\cite{Martinez_2017_CVPR}, the models are trained on 6 subjects and tested on the specific clips of the 5th subject.

{\bf CMU motion capture} (CMU Mocap). CMU Mocap consists of $5$ general classes of actions: `human interaction', `interaction with environment', `locomotion', `physical activities \& sports', and `situations \& scenarios', where each subject has $38$ joints and we preserve $26$ joints with non-zero exponential maps. Be consistent with~\cite{Li_2018_CVPR}, we  select $8$ detailed actions: `basketball', `basketball signal', `directing traffic', `jumping', `running', `soccer', `walking' and `washing window'. We evaluate our model with the same approach as we do for H3.6M.

\textbf{Model configuration.} 
We implement DMGNN with PyTorch 1.0 on one GTX-2080Ti GPU. We set $3$ scales, which contains body-joints, $10$ and $5$ body-components for both datasets. We use $4$ cascaded MGCUs, whose feature dimensions are $32$, $64$, $128$ and $256$, respectively. In the first two MGCUs, we use both SS-GCBs and CS-FBs to extract spatio-temporal features and fuse cross-scale features; In the last two MGCUs, we only use SS-GCBs. In the decoder, the dimension of the G-GRU is $256$, and we use a two-layer MLP for pose output. In training, we set the batch size $32$ and clip the gradients to a maximum $\ell_2$-norm of $0.5$; we use Adam optimizer~\cite{Kingma_iclr2015} with learning rate $0.0001$. All the hyper-parameters are selected with validation sets.

\textbf{Baseline methods.} 
We compare the proposed DMGNN with many recent works, which learned motion patterns from pose vectors, e.g. Res-sup.~\cite{Martinez_2017_CVPR}, 
CSM~\cite{Li_2018_CVPR}, 
TP-RNN~\cite{abs-1810-09676}, 
AGED~\cite{Gui_2018_ECCV}, 
and 
Imit-L~\cite{Wang_2019_ICCV}, 
or separated bodies e.g. 
Skel-TNet~\cite{AAAI_Guo}, 
and 
Traj-GCN~\cite{Mao_2019_ICCV}. 
We reproduce, Res-sup., CSM and Traj-GCN based on their released codes. We also employ a naive baseline, ZeroV~\cite{Martinez_2017_CVPR}, which sets all predictions to be the last observed pose at $t=0$. 

\subsection{Comparison to state-of-the-art methods}
To validate the proposed DMGNN, we show the prediction performance for both short-term and long-term motion prediction on Human 3.6M (H3.6M) and CMU Mocap. We quantitatively evaluate various methods by the mean angle error (MAE) between the generated motions and ground-truths in angle space. We also illustrate the predicted samples for qualitative evaluation.

\textbf{Short-term motion prediction.}
Short-term motion prediction aims to predict the future poses within 500 milliseconds. We compare DMGNN to state-of-the-art methods for predicting poses in 400 milliseconds on H3.6M dataset. We first test $4$ representative actions: `Walking', `Eating', `Smoking' and `Discussion'. Table~\ref{tab:pred_h36m_4} shows MAEs of DMGNN and some baselines. We also present the performance of several variants of DMGNN: we use fixed body-graphs in SS-GCBs (fixed $\mathbf{A}_{s}$); the common GRU without a graph (no G-GRU); or only the joint-scale ($S=1$) bodies. We see that, i) the complete DMGNN obtain the most precise prediction among all the variants; ii) compared to baselines, DMGNN has the lowest prediction MAEs on `Eating' and `Smoking', and obtains competitive results on `Walking' and `Discussion'. Table~\ref{tab:pred_h36m_11} compares the proposed DMGNN with some recent baselines on the remaining $11$ actions in H3.6M. We see that DMGNN achieves the best performance in most actions (also for average MAEs).

\textbf{Long-term motion prediction.}
Long-term motion prediction aims to predict the poses over 500 milliseconds, which is challenging due to the action variation and non-linearity movements.  Table~\ref{tab:longterm} presents the MAEs of various models for predicting $4$ actions and average MAEs across the $4$ actions in the future 560 ms and 1000 ms on H3.6M dataset. We see that DMGNN outperforms the competitors on actions `Eating', and `Discussion' at 560 ms, and obtains competitive  performances on other cases.

We also train our DMGNN for short-term and long-term prediction on $8$ classes of actions in CMU Mocap dataset.  Table~\ref{tab:pred_cmu} shows the MAEs across the future 1000 ms. We see that DMGNN significantly outperforms the state-of-the-art methods  on  actions `Basketball', `Basketball Signal', `Running' and `Walking' and  obtains competitive  performance on the other actions.

\begin{figure}[t]
    \centering
    \includegraphics[width=1\columnwidth]{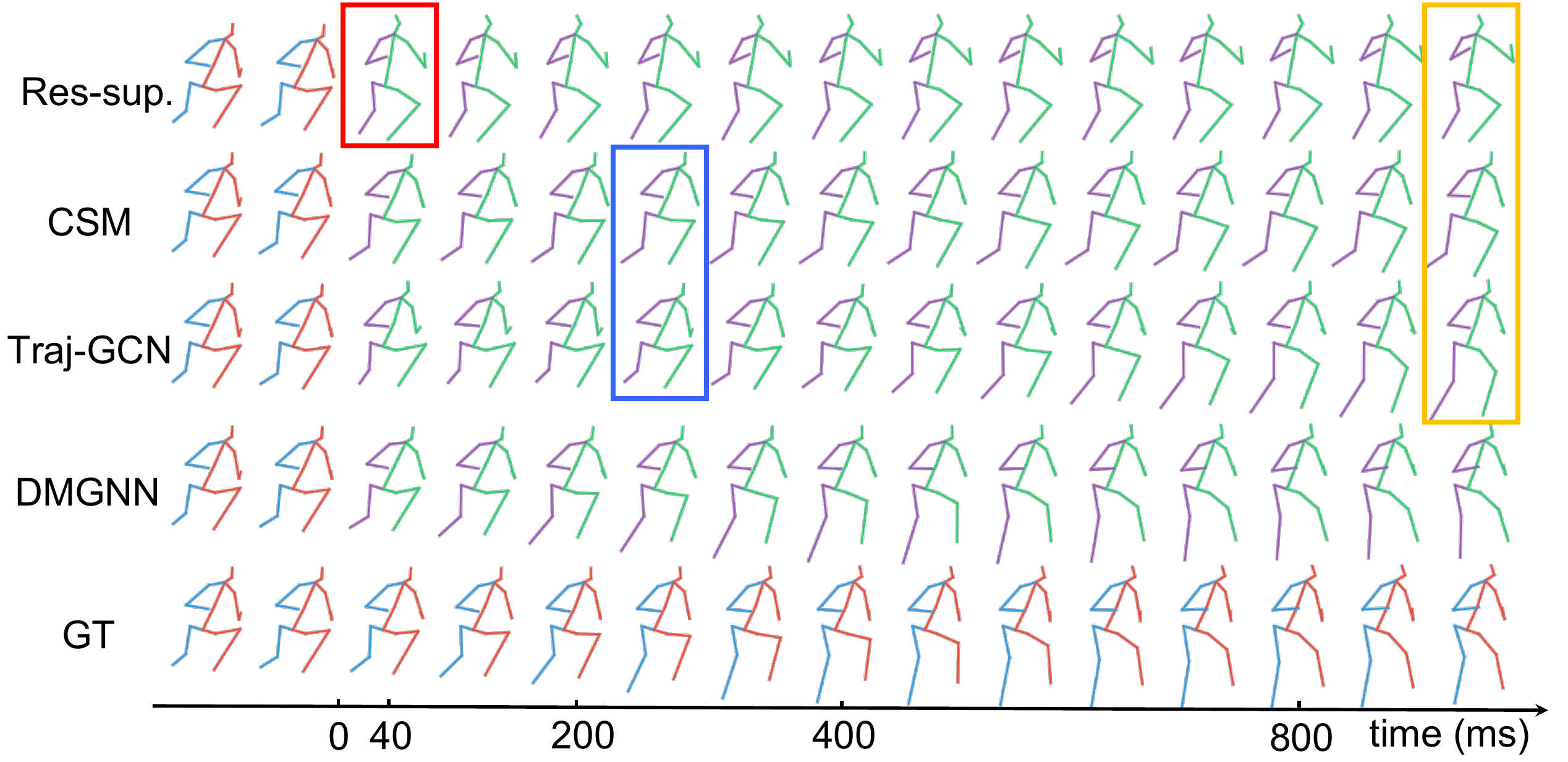}
    \caption{Qualitative comparison on the action `Taking Photo' of H3.6M for both short and long-term prediction.}
    \label{fig:sample_show}
    \vspace{-2mm}
\end{figure}

\textbf{Predicted sample visualization.} We compare the synthesized samples of DMGNN to those of Res-sup., CSM and Traj-GCN on H3.6M. Figure~\ref{fig:sample_show} illustrates the future poses of `Taking Photo' in 1000 ms with the frame interval of 80 ms. Comparing to baselines, we see that DMGNN completes the action accurately and reasonably, providing significantly better predictions. Res-sup. has large discontinuity between the last observed pose the first predicted one (red box); CSM and Traj-GCN have large errors after the 280th ms (blue box); three baselines give large posture errors in long-term (yellow box). We show more prediction images and videos in Appendix.

\textbf{Effectiveness and efficiency test.}  
We compare the running time costs of DMGNN to several latest models. Table~\ref{tab:run_time} presents the running time of different methods for short and long-term motion prediction on H3.6M dataset. We see that DMGNN achieves the shortest running time while generating future poses over both 400 or 1000 ms, compared with the other competitors~\cite{Martinez_2017_CVPR, Li_2018_CVPR, Mao_2019_ICCV}. DMGNN takes only $29.18$ ms to generate motions in 400 ms, indicating that DMGNN with multiscale graphs has efficient operations.

\begin{table}[t]
    \centering
    \caption{Average time cost comparison between DMGCNN with the latest models on H3.6M dataset.}
    \footnotesize
    \setlength{\tabcolsep}{8.1mm}{

        \begin{tabular}{l|c|c}
        \hline
        Model & \multicolumn{2}{|c}{Time cost (ms)}  \\
        \hline
        milisecond & 400 & 1000 \\
        \hline
        TP-RNN~\cite{abs-1810-09676} & 48.96 & 127.41\\
        Skel-TNet~\cite{AAAI_Guo} & 33.29 & 98.17 \\
        Traj-GCN~\cite{Mao_2019_ICCV} & 71.43 & 144.93\\
        DMGNN &  {\bf  29.18} & {\bf 86.04} \\
        \hline
        \end{tabular}}
        \vspace{-10pt}
    \label{tab:run_time}
\end{table}

\vspace{-1mm}
\subsection{Ablation study}
\vspace{-1mm}
We now investigate some crucial elements of DMGNN.

\textbf{Effects of multiple scales.}
\begin{table}[t]
    \centering
    \caption{Average MAEs of DMGNN with different scales for short-term prediction at different time stamps.}
    \footnotesize
    \setlength{\tabcolsep}{2.04mm}{

        \begin{tabular}{l|ccccc|cccc}
        \hline
        ~ & \multicolumn{5}{|c|}{Node numbers $M_s$} & \multicolumn{4}{|c}{MAEs} \\
        \hline
        Scales & 20 & 10 & 5 & 3 & 2 & 80 & 160 & 320 & 400\\
        \hline

        $1$ & \checkmark &  &  &  &  & 0.29 & 0.55 & 0.87 & 1.00 \\
        ${1,2}$ & \checkmark & \checkmark &  &  &  & {\bf 0.27} & 0.53 & 0.85 & 0.97 \\
        ${1,2,3}$ & \checkmark & \checkmark & \checkmark &  &  & {\bf 0.27} & {\bf 0.52} & {\bf 0.83} & {\bf 0.95} \\
        ${1,3}$ & \checkmark &  & \checkmark &  &  & 0.28 & 0.53 & 0.84 & {\bf 0.92} \\
        ${1,2,3,4}$ & \checkmark & \checkmark & \checkmark & \checkmark &  & 0.28 & 0.54 & 0.87 & 0.98 \\
        ${1,4}$ & \checkmark &  &  & \checkmark &  & 0.28 & 0.54 & 0.86 & 0.97 \\
        ${1,2,3,5}$ & \checkmark & \checkmark & \checkmark &  & \checkmark & 0.28 & 0.55 & 0.86 & 0.99 \\
        ${1,5}$ & \checkmark &  &  &  & \checkmark & 0.29 & 0.55 & 0.87 & 1.00 \\
        \hline
        \end{tabular}}
        \vspace{-10pt}
    \label{tab:scale_time}
\end{table}
To verify the proposed multiscale representation, we employ various scales in DMGNN for 3D skeleton-based motion prediction. Besides the three scales in our model, we introduce additional two scales: $s_4$, which represents a body as $M_{s_4}=3$ parts: left limbs, right limbs and torso, and $s_5$, which contains $M_{s_5}=2$ parts: upper body and lower body; see illustrations of $s_4$ and $s_5$ in Appendix. Table~\ref{tab:scale_time} presents the MAEs with various scales. We see that, when we combine $s_1$, $s_2$ and $s_3$, lowest prediction error is achieved. Notably, using two scales ($s_{1}, s_{2}$ or $s_{1}, s_{3}$) is significant better than using only $s_{1}$; but involving too abstract scales ($s_4$ or $s_5$) tends to hurt prediction.

\textbf{Effects of the number of MGCUs.}
\begin{table}[t]
    \centering
    \caption{MAEs and running times of DMGNN with different numbers of MGCUs for short and long-term prediction on H3.6M.}
    \footnotesize
    \setlength{\tabcolsep}{1.44mm}{
    
        \begin{tabular}{c|cccc|cc|cc}
        \hline
         ~ & \multicolumn{6}{|c|}{MAE at different time stamps (ms)} &\multicolumn{2}{|c}{running time (ms)}\\
        \hline
         MGCUs & 80&160&320&400&560&1000 & 400 & 1000\\
        \hline

        1 & 0.30 & 0.56 & 0.87 & 1.02 & 1.25 & 1.52 & 27.42 & 83.01\\
        2 & 0.29 & 0.53 & 0.85 & 0.99 & 1.20 & 1.52 & 27.89 & 83.95\\
        3 & {\bf 0.27} & 0.54 & {\bf 0.83} & {\bf 0.95} & {1.18} & {1.49} & 28.34 & 84.89\\
        4 & {\bf 0.27} & {\bf 0.52} & {\bf 0.83} & {\bf 0.95} & {\bf 1.16} & {\bf 1.48} & 29.18 & 86.04\\
        5 & 0.28 & 0.55 & {\bf 0.83} & {0.96} & 1.17 & {1.51} & 30.37 & 88.39\\
        6 & {0.29} & {0.54} & 0.84 & {0.98} & 1.19 & 1.54 & 31.55 & 91.15\\
        \hline
        \end{tabular}}
        \vspace{-10pt}
    \label{tab:layer}
\end{table}
To validate the effects of multiple MGCUs in the encoder, we tune the numbers of MGCUs from $1$ to $6$ and show the prediction errors and running time costs for short and long-term prediction on H3.6M, which are presented in Table~\ref{tab:layer}. We see that, when we adopt $1$ to $4$ MGCUs, the prediction MAEs fall and time costs rise continuously; when we use $5$ or $6$ MGCUs, the prediction errors are stably low, but the time costs rise higher. Therefore, we select to use $4$ MGCUs, resulting in precise prediction and high running efficiency.

\textbf{Effects of CS-FBs.}
\begin{table}[t]
    \centering
    \caption{Average MAEs of DMGNN with different numbers of CS-FBs and feature aggregators over  400 ms on H3.6M.}
    \footnotesize
    \setlength{\tabcolsep}{3.5mm}{
        \begin{tabular}{l|ccc|c}
        \hline
        ~ & \multicolumn{4}{|c}{Average MAE across 400 ms} \\
        \hline
        CS-FB numbers & 1 & 2 & 3 & 0  \\
        \hline
        without relative & 0.623 & 0.622 & 0.618 & \multirow{2}{*}{0.630} \\
        with relative & 0.618 & {\bf 0.613} & 0.616 &  \\
        \hline
        \end{tabular}}
        \vspace{-10pt}
    \label{tab:fusion_mechanism}
\end{table}
Here, we evaluate 1) the effectiveness of using relative features during cross-scale graph inference in CS-FBs; 2) different numbers of CS-FBs in a sequence of $4$ MGCUs.  For $0$ CS-FB, the model only fuses all scales at the end of the encoder. Table~\ref{tab:fusion_mechanism} presents the average MAEs with different CS-FBs and relative-feature mechanisms across 400 ms on H3.6M. We see that 1) using relative features leads to lower MAEs, validating the effectiveness of such augmented features; 2) $2$ CS-FBs leads to the best prediction performance. The intuition is that 0 or 1 CS-FB fuse insufficiently and 3 CS-FBs tend to fuse redundant information to confuse the model. 

\begin{figure}[t]
    \centering
    \includegraphics[width=0.8\columnwidth]{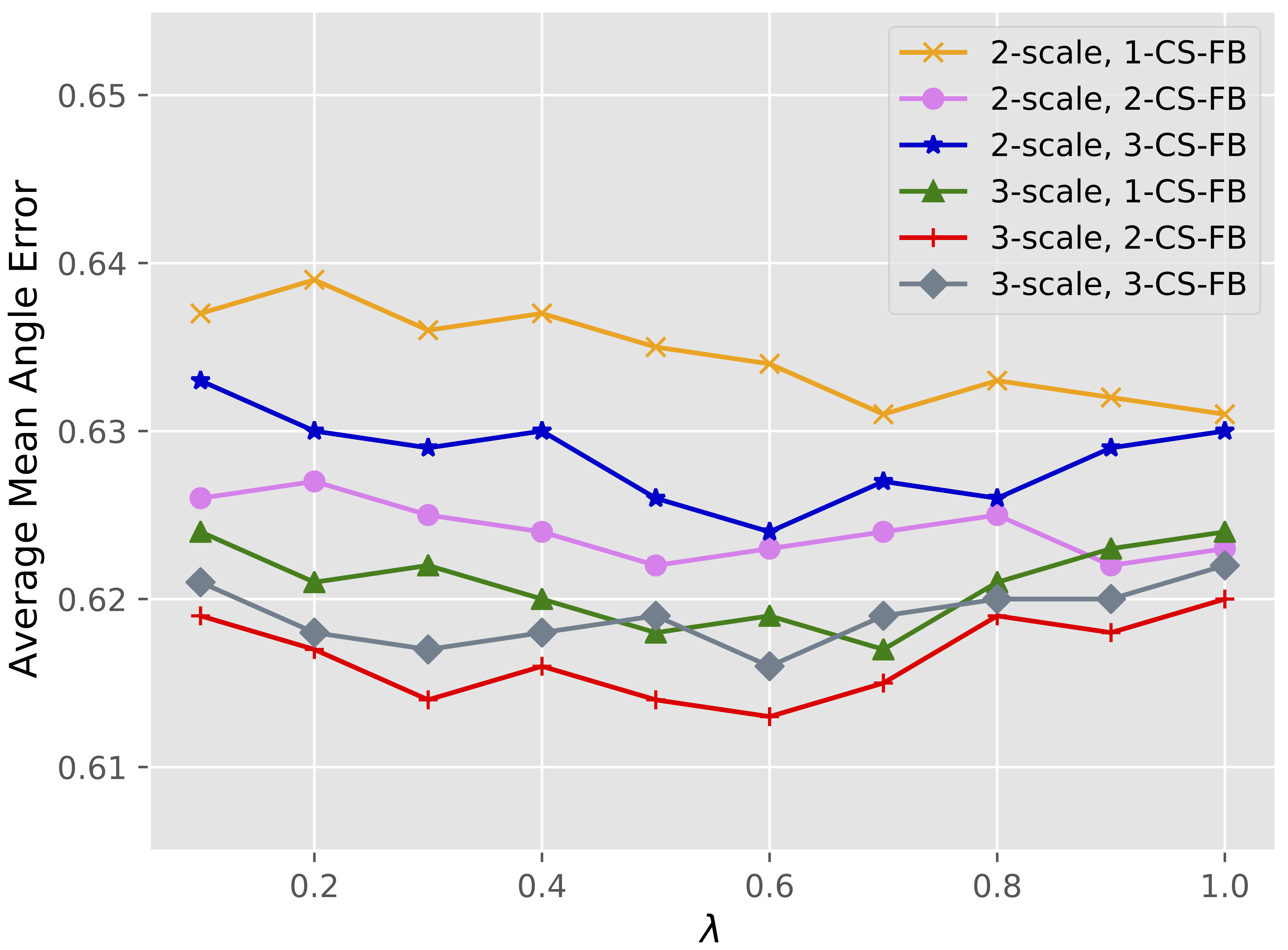}
    \caption{Average MAEs of DMGNN variants with different final fusion coefficient $\lambda$ for short-term motion prediction.}
    \label{fig:mae-lamda}
    \vspace{-5pt}
\end{figure}

\textbf{Effect of $\lambda$ in final fusion.} 
The hyper-parameter $\lambda$ in the final fusion~\eqref{eq:final_fusion} balances the influence between joint-scale and more abstract scales. Figure~\ref{fig:mae-lamda} illustrates the average MAE with different body scales and CS-FBs for short-term prediction on H3.6M. We see that the performance reach its best when we use $3$ scales, $2$ hierarchical CS-FBs and $\lambda=0.6$, even though it is robust to the change of $\lambda$.

    

\textbf{Effect of high-order motion differences.}
\begin{table}[t]
    \centering
    \caption{Average MAEs for different orders of motion differences.}
    \footnotesize
    \setlength{\tabcolsep}{4.4mm}{
    
        \begin{tabular}{c|cccc}
        \hline
          & \multicolumn{4}{|c}{MAE at different time stamps (ms)} \\
        \hline
        Difference Order & 80&160&320&400 \\
        \hline

        $\beta = 0$ & 0.34 & 0.60 & 0.86 & 1.01 \\
        $\beta = 0,1$ & 0.28 & 0.54 & {\bf 0.83} & 0.97 \\
        $\beta = 0,1,2$ & {\bf 0.27} & {\bf 0.52} & {\bf 0.83} & {\bf 0.95} \\
        \hline
        \end{tabular}}
        \vspace{-5pt}
    \label{tab:difference}
\end{table}
We study the effects of various orders of motion differences fed into the encoder and decoder of our model. We evaluate DMGNN with combinations of $0,1,2$-orders of pose differences. Table~\ref{tab:difference} presents the MAEs of DMGNN with various input differences for short-term motion prediction. We see that the proposed DMGNN obtains the lowest MAEs when it adopts the $0,1,2$-orders of motion differences. This indicates that high-order differences improve the prediction performance significantly.

\subsection{Analysis of category-agnostic property}
\vspace{-1mm}
Here we validate that DMGNN can learn discriminative motion features for category-agnostic prediction. 

\begin{figure}[t]
    \small
    \centering
    \footnotesize
    \includegraphics[width=0.9\columnwidth]{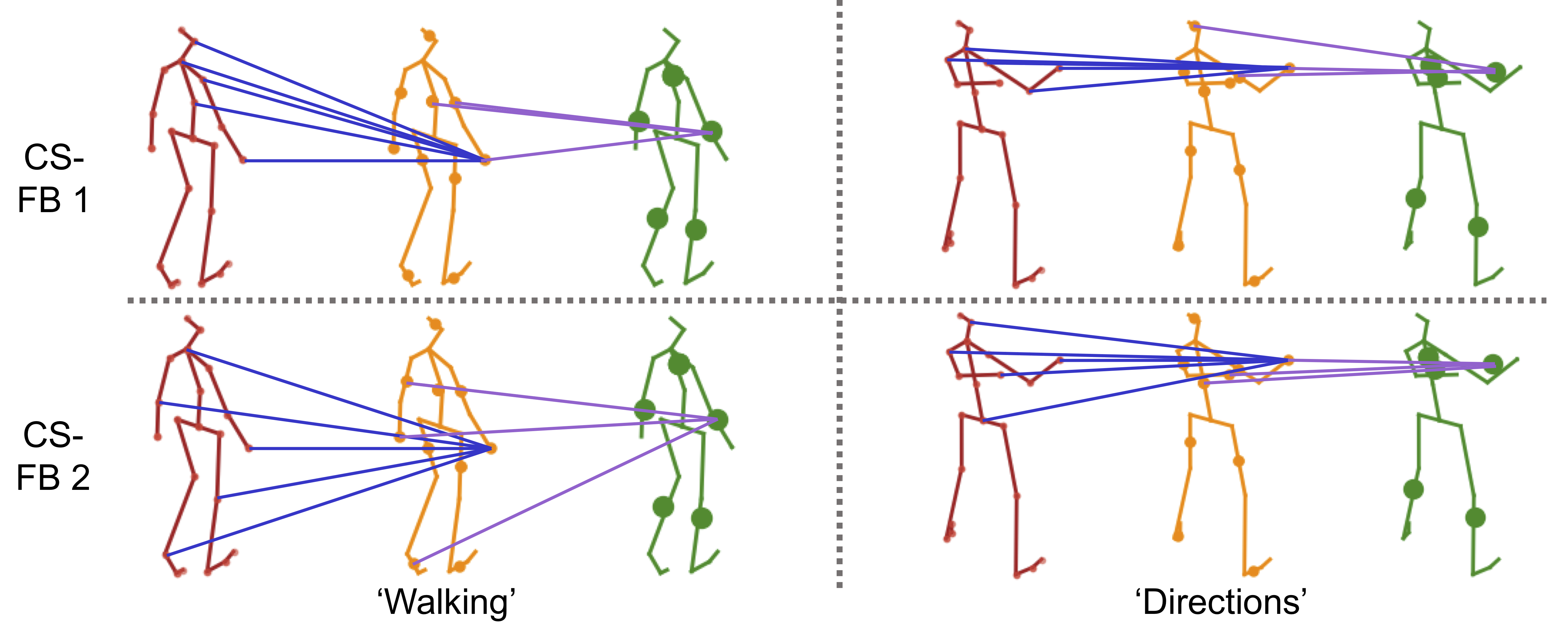}
    \caption{The learned dynamic cross-scale graphs on two CS-FBs for two actions: `Walking' and `Directions' in H3.6M.}
    \label{fig:learned_graph}
    \vspace{-5pt}
\end{figure}

We first visualize the learned cross-scale graphs for different actions to test the discriminative power. Figure~\ref{fig:learned_graph} shows the graphs in two CS-FBs on `Walking' and `Directions' in H3.6M. For each action, we show some strong relations from detailed scales to the right arms in coarse scales. We see that i) for each action, the CS-FBs capture diverse ranges of a human body: the graph in the first CS-FB focuses on nearby body-components; the second CS-FB captures more global and action-related effects; i.e. hands and feet affects arms during walking; and ii) the cross-scale graphs are different for various actions, especially in the second CS-FB, capturing distinct patterns.

\begin{table}[t]
    \centering
    \caption{Classification accuracies on cross-scale graphs and motion features of DMGNN and other methods on H3.6M.}
    \footnotesize
    \setlength{\tabcolsep}{0.69mm}{
    \begin{tabular}{c|ccccc}
    \hline
       Methods & On CS-FB 1 & On CS-FB 2 & On $\mathbf{H}$ & Res-sup.~\cite{Martinez_2017_CVPR} & TP-RNN~\cite{abs-1810-09676} \\
    \hline
       Accuracy & 28.6\%  & 40.1\%  & 45.7\% & 22.6\% & 24.4\% \\
    \hline
    \end{tabular}}
    \label{tab:graph_classification}
    \vspace{-10pt}
\end{table}

We next conduct action classification on the intermediate representations to test the discriminative power. We isolatedly train a two-layer MLP to classify each dynamic cross-scale graph. We also classify the outputs from the encoders of DMGNN, Res-sup. (class-aware) and TP-RNN (class-agnostic). Table~\ref{tab:graph_classification} presents the average classification accuracies on $15$ categories of actions. We see that the cross-scale graph in the second CS-FB is more informative than the one in the first CS-FB for action recognition. Comparing to baselines, DMGNN obtains the highest the classification accuracies on encoder representation, indicating that DMGNN captures discriminative information for class-agnostic prediction. 

\section{Conclusion}
We build dynamic mutiscale graphs to represent a human body and propose dynamic multiscale graph neural networks (DMGNN) with an encoder-decoder framework for 3D skeleton-based human motion prediction. In the encoder, We develop multiscale graph computational units (MGCU) to extract features; in the decoder, we develop a graph-based GRU (G-GRU) for pose generation. The results show that the proposed model outperforms most state-of-the-art methods for both short and long-trem prediction in terms of both effectiveness and efficiency.

\textbf{Acknowledgement:}
This work is supported by the National Key Research and Development Program of China (No. 2019YFB1804304), SHEITC (No. 2018-RGZN-02046), NSFC (No. 61521062), 111 plan (No. B07022),  and STCSM (No. 18DZ2270700).

\newpage
{\small
\bibliographystyle{ieee_fullname}
\bibliography{egbib}
}

\newpage

\section{Detailed Architecture}
Here we show the detailed structure of the proposed DMGNN. We first show the structure of the encoder, including the single-scale graph convolution block (SS-GCB) and cross-scale fusion block (CS-FB). We then show the structure of the decoder, including the graph-based gated recurrent unit (G-GRU).

\subsection{Encoder}
\textbf{Single-scale graph convolution block (SS-GCB).} SS-GCB consists of a graph convolution and a temporal convolution. Table~\ref{tab:SS-GCB} presents the structures of four cascaded SS-GCB at scale $s$ in the encoder of DMGNN.
\begin{table}[htb]
    \centering
    \caption{The structure of four SS-GCBs at scale $s$ in the encoder.}
    \small
    \setlength{\tabcolsep}{0.8mm}{
    \begin{tabular}{c|c|c|c}
    \hline
        Idx & Shape \& Operations& Feature & Remarks \\ \hline
        \multirow{4}{*}{1} 
                          & $[32, 3, 1, 1]\times2$ 
                          & \multirow{2}{*}{$[32, 32, M_s, 49]$} 
                          & \multirow{2}{*}{graph conv}\\
        ~                 & -bn-relu 
                          & ~\\ 
                          \cline{2-4}
        ~                 & $[32, 32, 5, 1]$, stride=1
                          & \multirow{2}{*}{$[32, 32, M_s, 49]$}
                          & \multirow{2}{*}{ temporal conv} \\ 
        ~                 & bn-dropout-relu 
                          & ~ \\ 
                          \hline
        
        \multirow{4}{*}{2} 
                          & $[64, 32, 1, 1]\times2$ 
                          & \multirow{2}{*}{$[32, 64, M_s, 49]$}
                          & \multirow{2}{*}{graph conv}\\
        ~                 & -bn-relu 
                          & ~\\ 
                          \cline{2-4}
        ~                 & $[64, 64, 5, 1]$, stride=2 
                          & \multirow{2}{*}{$[32, 64, M_s, 25]$}
                          & \multirow{2}{*}{temporal conv} \\ 
        ~                 & bn-dropout-relu 
                          & ~ \\ 
                          \hline
        
        \multirow{4}{*}{3} 
                          & $[128, 64, 1, 1]\times2$ 
                          & \multirow{2}{*}{$[32, 128, M_s, 25]$}
                          & \multirow{2}{*}{graph conv}\\
        ~                 & -bn-relu 
                          & ~\\ 
                          \cline{2-4}
        ~                 & $[128, 128, 5, 1]$, stride=2 
                          & \multirow{2}{*}{$[32, 128, M_s, 13]$}
                          & \multirow{2}{*}{temporal conv} \\ 
        ~                 & bn-dropout-relu 
                          & ~ \\ 
                          \hline
        
        \multirow{4}{*}{4} 
                          & $[256, 128, 1, 1]\times2$ 
                          & \multirow{2}{*}{$[32, 256, M_s, 13]$}
                          & \multirow{2}{*}{graph conv}\\
        ~                 & -bn-relu 
                          & ~\\ \cline{2-4}
        ~                 & $[256, 256, 5, 1]$, stride=2 
                          & \multirow{2}{*}{$[32, 256, M_s, 7]$}
                          & \multirow{2}{*}{temporal conv} \\ 
        ~                 & bn-dropout-relu 
                          & ~ \\ \hline
        
    \end{tabular}}
    \label{tab:SS-GCB}
\end{table}
We see that we use four SS-GCBs to extract spatio-temporal motion features. In each SS-GCB, we employ ReLU, batch normalization, and dropout operations. We use stride 2 to downsample along the temporal dimension.

\textbf{Cross-scale fusion block (CS-FB)}
We use CS-FB to fuse multiscale features. Table~\ref{tab:CS-FB} presents the structure of the first CS-FB to fuse the feature from $s_1$ to $s_2$.
\begin{table}[htb]
    \centering
    \caption{The structure of the first CS-FB from $s_1$ to $s_2$.}
    \small
    \setlength{\tabcolsep}{2.4mm}{
    \begin{tabular}{c|c}
    \hline
        Step & Shape \& Operations \\ \hline
        1 & temporal conv: $[32, 32, 5, 1]$, stride=2; vectorize \\
        \hline
        \multirow{4}{*}{2} & for both $f_{s_1}$ and $f_{s_2}$: 800-256-relu \\
        ~ & -dropout-256-relu-bn; Sum \\ \cline{2-2}
        ~ & for both $g_{s_1}$ and $g_{s_2}$: 512-256-relu \\
        ~ & -dropout-256-relu-bn \\ \hline
        
        3 & Computing (2e) in paper \\ \hline
    \end{tabular}}
    \label{tab:CS-FB}
\end{table}
We first use a temporal convolution to shrink the temporal dimension and obtain a compact feature vector for each body-component; we then use four MLPs to learn the feature embeddings for two body-scales, respectively;  we finally calculate the inner product of these two embeddings and employ a softmax to calculate the corresponding edge weight in a cross-scale graph. 

\textbf{Total architecture}
In summary, we show the total architecture of the encoder, which combine SS-GCBs at multiple scales and CS-FB across scales. Table~\ref{tab:encoder} presents the structure of the encoder.
\begin{table}[ht]
    \centering
    \caption{The structure of the encoder.}
    \small
    \setlength{\tabcolsep}{1.8mm}{
    \begin{tabular}{c|c|c|c|c|c|c}
        \hline
        MGCU &
        \multicolumn{6}{c}{Initialize three scales} \\ 
        \hline
        \multirow{2}{*}{1} &
        \multicolumn{2}{c|}{SS-GCB 1 at $s_1$} & 
        \multicolumn{2}{|c|}{SS-GCB 1 at $s_2$} & \multicolumn{2}{|c}{SS-GCB 1 at $s_3$} \\
        \cline{2-7}
        ~ &
        \multicolumn{6}{c}{CS-FB 1 between $s_1$\&$s_2$ and $s_2$\&$s_3$} \\
        \hline
        \multirow{2}{*}{2} &
        \multicolumn{2}{c|}{SS-GCB 2 at $s_1$} & 
        \multicolumn{2}{|c|}{SS-GCB 2 at $s_2$} & \multicolumn{2}{|c}{SS-GCB 2 at $s_3$} \\
        \cline{2-7}
        ~ &
        \multicolumn{6}{c}{CS-FB 2 between $s_1$\&$s_2$ and $s_2$\&$s_3$} \\
        \hline
        3 &
        \multicolumn{2}{c|}{SS-GCB 3 at $s_1$} & 
        \multicolumn{2}{|c|}{SS-GCB 3 at $s_2$} & \multicolumn{2}{|c}{SS-GCB 3 at $s_3$} \\
        \hline
        4 &
        \multicolumn{2}{c|}{SS-GCB 4 at $s_1$} & 
        \multicolumn{2}{|c|}{SS-GCB 4 at $s_2$} & \multicolumn{2}{|c}{SS-GCB 4 at $s_3$} \\
        \hline
        ~ &
        \multicolumn{6}{c}{Weighted sum} \\ \hline
        ~ &
        \multicolumn{6}{c}{A final SS-GCB at $s_1$} \\ \hline
        ~ &
        \multicolumn{6}{c}{Temporal average pooling} \\
        \hline
    \end{tabular}}
    \label{tab:encoder}
\end{table}
We see that we use four MGCUs, where the first two MGCUs use SS-GCBs and CS-FBs to learn the features from multiscale bodies and the last two MGCUs only use SS-GCB to extract features.

\subsection{Decoder}
\textbf{Graph-based Gated Recurrent Unit (G-GRU)}
G-GRU is one of the key components in the proposed decoder for synthesizing precise and reasonable future poses. Table~\ref{tab:G-GRU} presents the structure of the G-GRU at time stamp $t$.
\begin{table}[ht]
    \centering
    \caption{The structure of the G-GRU in the decoder at time $t$.}
    \small
    \setlength{\tabcolsep}{1.8mm}{
    \begin{tabular}{c|c}
        \hline
        Variables & Operations \\
        \hline
        \multirow{3}{*}{$\mathbf{r}^{(t)}$} & input: $\mathbf{H}^{(t)}$, $\mathbf{I}^{(t)}$; $r_{\rm in}$: $9\to256$ \\
        ~ & graph conv: $256\to256$; $r_{\rm hid}$: $256\to256$ \\
        ~ & sum and sigmoid \\
        \hline
        \multirow{3}{*}{$\mathbf{u}^{(t)}$} & input: $\mathbf{H}^{(t)}$, $\mathbf{I}^{(t)}$; $u_{\rm in}$: $9\to256$ \\
        ~ & graph conv: $256\to256$; $u_{\rm hid}$: $256\to256$ \\
        ~ & sum and sigmoid \\
        \hline
        \multirow{4}{*}{$\mathbf{c}^{(t)}$} & input: $\mathbf{H}^{(t)}$, $\mathbf{I}^{(t)}$; $c_{\rm in}$: $9\to256$ \\
        ~ & graph conv: $256\to256$; $c_{\rm hid}$: $256\to256$ \\
        ~ & element-wise product of $c_{\rm hid}$ and $\mathbf{r}^{(t)}$ \\
        ~ & sum and tanh \\
        \hline
        $\mathbf{H}^{(t+1)}$ & $\mathbf{u}^{(t)}\odot\mathbf{H}^{(t)}+(1-\mathbf{u}^{(t)})\odot\mathbf{c}^{(t)}$\\
        \hline 
    \end{tabular}}
    \label{tab:G-GRU}
\end{table}
We see that we take the historical motion state and the online 3D skeleton-based information as inputs and introduce the graph convolution to propagate the motion information to produce the motion state at the next frame. The hidden dimension of the G-GRU is 256.

\textbf{Total architecture}
Here, we show the total architecture of the decoder, which combines the proposed G-GRU and an MLP-formed output function. Table~\ref{tab:decoder} presents the structure the decoder at time stamp $t$.
\begin{table}[ht]
    \centering
    \caption{The structure of the decoder at time $t$.}
    \small
    \setlength{\tabcolsep}{1.8mm}{
    \begin{tabular}{c|c}
        \hline
        ~ & Operations \\
        \hline
        Inputs & $\mathbf{H}^{(t)}$, 
                 $\mathbf{I}^{(t)}=[\widehat{\mathbf{X}}^{(t)},
                                    \Delta^1\widehat{\mathbf{X}}^{(t)},
                                    \Delta^2\widehat{\mathbf{X}}^{(t)}]$ \\
        \hline
        G-GRU & $\mathbf{H}^{(t+1)}=\textrm{G-GRU}(\mathbf{I}^{(t)}, \mathbf{H}^{(t)})$, $9,256\to256$ \\
        \hline
        $f_{\rm pred}$ & $f_{\rm pred}(\mathbf{H}^{(t+1)})$, $256\to256\to3$\\
        \hline
        $\widehat{\mathbf{X}}^{(t+1)}$ & $\widehat{\mathbf{X}}^{(t+1)}=\widehat{\mathbf{X}}^{(t)}+
                                      f_{\rm pred}(\mathbf{H}^{(t+1)})$ \\
        \hline
    \end{tabular}}
    \label{tab:decoder}
\end{table}
We see that, given the hidden motion state and current input information, we use a G-GRU and an MLP-formed output function $f_{\rm pred}$ to model the displacement of motions between two consecutive frames, and we emply residual connections to obtain the estimated poses. The hidden dimensions are 256.

\section{Quantitative Comparison with more Baselines}
In our paper submission, we only compare DMGNN to several state-of-the-art works, while many other methods has been developed. Here we compare DMGNN to as many previous methods as possible. Table~\ref{tab:pred_h36m_4} presents the MAE of many methods for short-term motion prediction on 4 representative actions of Human 3.6M
\begin{table*}[htb]
    \centering
    \caption{Mean angle errors (MAE) of different methods for short-term prediction on 4 representative actions of H3.6M.}
    \footnotesize
    \setlength{\tabcolsep}{2mm}{

        \begin{tabular}{c|cccc|cccc|cccc|cccc}
        \hline
        Motion & \multicolumn{4}{|c|}{Walking} & \multicolumn{4}{|c|}{Eating} & \multicolumn{4}{|c|}{Smoking} & \multicolumn{4}{|c}{Discussion}\\
        \hline
        milliseconds & 80&160&320&400 & 80&160&320&400 & 80&160&320&400 & 80&160&320&400 \\
        \hline
        ZeroV~\cite{Martinez_2017_CVPR} & 0.39 & 0.68 & 0.99 & 1.15 & 0.27 & 0.48 & 0.73 & 0.86 & 0.26 & 0.48 & 0.97 & 0.95 & 0.31 & 0.67 & 0.94 & 1.04 \\
        ERD~\cite{Fragkiadaki_2015_ICCV} & 0.93 & 1.18 & 1.59 & 1.78 & 1.27 & 1.45 & 1.66 & 1.80 & 1.66 & 1.95 & 2.35 & 2.42 & 0.31 & 0.67 & 0.94 & 1.04 \\
        LSTM-3R~\cite{Fragkiadaki_2015_ICCV} & 0.977 & 1.00 & 1.29 & 1.47 & 0.89 & 1.09 & 1.35 & 1.46 & 1.34 & 1.65 & 2.04 & 2.16 & 1.88 & 2.12 & 2.25 & 2.23\\
        SRNN~\cite{Jain_2016_CVPR} & 0.81 & 0.94 & 1.16 & 1.30 & 0.97 & 1.14 & 1.35 & 1.46 & 1.45 & 1.68 & 1.94 & 2.08 & 1.22 & 1.49 & 1.83 & 1.93 \\
        DropAE~\cite{GhoshSAH17} & 1.00 & 1.11 & 1.39 & / & 1.31 & 1.49 & 1.86 & / & 0.92 & 1.03 & 1.15 & / & 1.11 & 1.20 & 1.38 & / \\
        Res-sup.~\cite{Martinez_2017_CVPR} & 0.27 & 0.46 & 0.67 & 0.75 & 0.23 & 0.37 & 0.59 & 0.73 & 0.32 & 0.59 & 1.01 & 1.10 & 0.30 & 0.67 & 0.98 & 1.06 \\
        CSM~\cite{Li_2018_CVPR} & 0.33 & 0.54 & 0.68 & 0.73 & 0.22 & 0.36 & 0.58 & 0.71 & 0.26 & 0.49 & 0.96 & 0.92 & 0.32 & 0.67 & 0.94 & 1.01 \\
        TP-RNN~\cite{abs-1810-09676} & 0.25 & 0.41 & 0.58 & 0.65 & 0.20 & 0.33 & 0.53 & 0.67 & 0.26 & 0.47 & 0.88 & 0.90 & 0.30 & 0.66 & 0.96 & 1.04 \\
        QuaterNet~\cite{quater} & 0.21 & 0.34 & 0.56 & 0.62 & 0.20 & 0.35 & 0.58 & 0.70 & 0.25 & 0.47 & 0.93 & 0.90 & 0.26 & 0.60 & 0.85 & 0.93\\
        AGED~\cite{Gui_2018_ECCV} & 0.21 & 0.35 & 0.55 & 0.64 & 0.18 & 0.28 & 0.50 & 0.63 & 0.27 & 0.43 & 0.81 & 0.83 & 0.26 & {0.56} & {0.77} & {0.84} \\
        Skel-TNet~\cite{AAAI_Guo} & 0.31 & 0.50 & 0.69 & 0.76 & 0.20 & 0.31 & 0.53 & 0.69 & 0.25 & 0.50 & 0.93 & 0.89 & 0.30 & 0.64 & 0.89 & 0.98 \\
        BiHMP-GAN~\cite{AAAI_Kundu} & 0.33 & 0.52 & 0.63 & 0.67 & 0.20 & 0.33 & 0.54 & 0.70 & 0.26 & 0.50 & 0.91 & 0.86 & 0.33 & 0.65 & 0.91 & 0.95 \\
        VGRU-r1~\cite{Gopalakrishnan_2019_CVPR} & 0.34 & 0.47 & 0.64 & 0.72 & 0.27 & 0.40 & 0.64 & 0.79 & 0.36 & 0.61 & 0.85 & 0.92 & 0.46 & 0.82 & 0.95 &1.21 \\
        HMR~\cite{Liu_2019_CVPR} & 0.23 & 0.35 & 0.56 & 0.65 & 0.21 & 0.32 & 0.55 & 0.67 & 0.26 & 0.47 & 0.90 & 0.89 & 0.29 & 0.55 & 0.83 & 0.94 \\
        Imit-L~\cite{Wang_2019_ICCV} & 0.21 & 0.34 & 0.53 & 0.59 & 0.17 & 0.30 & 0.52 & 0.65 & 0.23 & 0.44 & 0.87 & 0.85 & 0.23 & 0.56 & 0.82 & 0.91 \\
        Traj-GCN~\cite{Mao_2019_ICCV} & {\bf 0.18} & {0.32} & {\bf 0.49} & {\bf 0.56} & {\bf 0.17} & 0.31 & 0.52 & 0.62 & 0.22 & 0.41 & 0.84 & 0.79 & {\bf 0.20} & {\bf 0.51} & {\bf 0.79} & {\bf 0.87}\\
        \hline
        DMGNN & {\bf 0.18} & {\bf 0.31} & {\bf 0.49} & {0.58} & {\bf 0.17} & {\bf 0.30} & {\bf 0.49} & {\bf 0.59} & {\bf 0.21} & {\bf 0.39} & {\bf 0.81} & {\bf 0.77} & {0.26} & {0.65} & {0.92} & {0.99}\\
        \hline
        \end{tabular}}
    \label{tab:pred_h36m_4}
\end{table*}
We see that, the proposed DMGNN outperforms the state-of-the-art methods on most actions. Notably, we have cited all of baselines presented in Table ~\ref{tab:pred_h36m_4} in our paper submission.

\section{Coarser Body-scales in Ablation Studies}
In the first experiment of ablation studies (`effects of multiple scales'), we initialize two coarser body-scales ($s_4$ and $s_5$) besides the effective three scales ($s_1$, $s_2$ and $s_3$) that used in our DMGNN. Here we present $s_4$ and $s_5$ in details.

\begin{figure}[t]
    \centering
    \includegraphics[width=0.9\columnwidth]{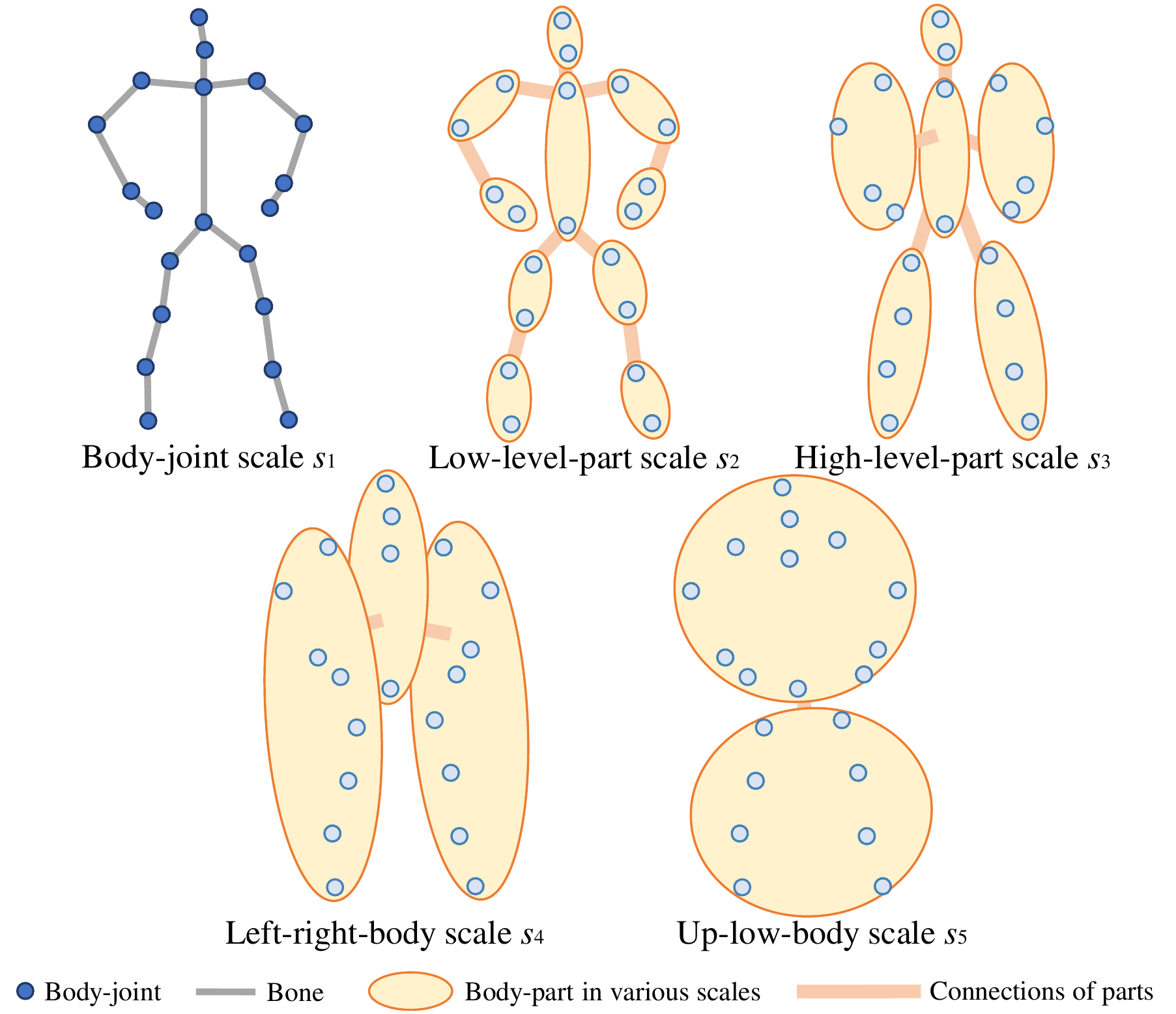}
    \caption{Three body scales on Human 3.6M. In body-joint scale, we consider 20 joints with non-zero exponential maps~\cite{Huynh2009}; In $s_4$ and $s_5$, we consider 3 and 2 parts, respectively.}
    \label{fig:multiscale_graphs_appendix}
\end{figure}

To initialize $s_4$, we average the input features of three body-components: left-body, head-and-torso, and right-body as the nodes of corresponding body-graph. We build two initial edges to respectively connect head-and-torso with left-body and right-body. To initialize $s_5$, we average the input features of two body-components: upper-body and lower-body as the graph nodes. We build an edge between these two body-components. Figure~\ref{fig:multiscale_graphs_appendix} illustrates the two coarser body-scales as well as the body-joint scale on Human 3.6M~\cite{6682899}. We name $s_4$ as `Left-right-body scale' and name $s_5$ as `Up-low-body scale'.  

\section{Effects of Numbers and Positions of CS-FBs}
In our DMGNN, we employ CS-FBs with aggregating  relative features at different MGCUs to fuse various levels of motion features across different scales; see Equation (2a) in the submission. Here we further investigate the effects of numbers and positions of CS-FBs at cascaded MGCUs. In the four MGCUs, we use one to four CS-FBs at different MGCUs, and we obtain the average prediction MAEs of different model variants. 

Table~\ref{tab:CS-FB_location} presents the average MAEs of DMGNN with different numbers of CS-FBs at different MGCUs on H3.6M for short-term motion prediction. We also compare the performance of CS-FBs with or without aggregating relative information from all the body-components (`with relative' or `without relative'). We denote the numbers of CS-FBs at the column `Number' and denote the CS-FB positions as MGCU indices at column `Position'. 
\begin{table}[htb]
    \centering
    \caption{Average MAEs of DMGNN with different numbers of CS-FBs at different MGCUs on H3.6M across 400 ms.}
    \small
    \setlength{\tabcolsep}{0.8mm}{
    \begin{tabular}{c|c|c|c}
    \hline
        Number & Position & MAE (without relative) & MAE (with relative) \\ \hline
        \multirow{4}{*}{1} & 1 & 0.621 & 0.621 \\ \cline{2-4}
                           & 2 & 0.620 & 0.618 \\ \cline{2-4}
                           & 3 & 0.620 & 0.616 \\ \cline{2-4}
                           & 4 & 0.622 & 0.619 \\ \hline
        \multirow{6}{*}{2} & 1,2 & 0.620 & {\bf 0.613} \\ \cline{2-4}
                           & 1,3 & 0.619 & 0.614 \\ \cline{2-4}
                           & 1,4 & 0.621 & 0.615 \\ \cline{2-4}
                           & 2,3 & 0.622 & 0.616 \\ \cline{2-4}
                           & 2,4 & 0.622 & 0.617 \\ \cline{2-4}
                           & 3,4 & 0.625 & 0.620 \\ \hline
        \multirow{4}{*}{3} & 1,2,3 & 0.622 & 0.616 \\ \cline{2-4}
                           & 1,2,4 & 0.623 & 0.619 \\ \cline{2-4}
                           & 1,3,4 & 0.624 & 0.622 \\ \cline{2-4}
                           & 2,3,4 & 0.625 & 0.622 \\ \hline
        4 & 1,2,3,4 & 0.622 & 0.619 \\ \hline
        0 & / & \multicolumn{2}{|c}{0.630} \\ \hline
    \end{tabular}}
    \label{tab:CS-FB_location}
\end{table}
We see that 1) when we aggregate global relative information to in the CS-FB, we obtain lower MAEs than the module without relative information aggregation; 2) when we use two CS-FBs with relative information aggregation at the 1st and 2nd MGCUs, DMGNN produces the most precise predictions across different model variants; 3) fusing multiscale features at first few MGCUs outperforms fusing at last ones. The reason behind could be, if we use only one CS-FB, we cannot fuse rich features for comprehensive pattern learning; if we use too many CS-FBs, the capacity of the network become much larger, leading to overfitting.

\section{More Generated Motion Samples}
To further demonstrate the effectiveness of the ASGNN, we illustrate more predicted samples on both Human 3.6M~\cite{6682899} and CMU Mocap~\footnote{http://mocap.cs.cmu.edu/} dataset.

\subsection{Human 3.6M Dataset}
We first illustrate two generated motions of the actions of `Posing' and `Waiting' on Human 3.6 dataset (H3.6). We compare the DMGNN with three models: Res-sup.~\cite{Martinez_2017_CVPR}, CSM~\cite{Li_2018_CVPR} and Traj-GCN~\cite{Mao_2019_ICCV}. 

\begin{figure}[t]
    \centering
    \includegraphics[width=1\columnwidth]{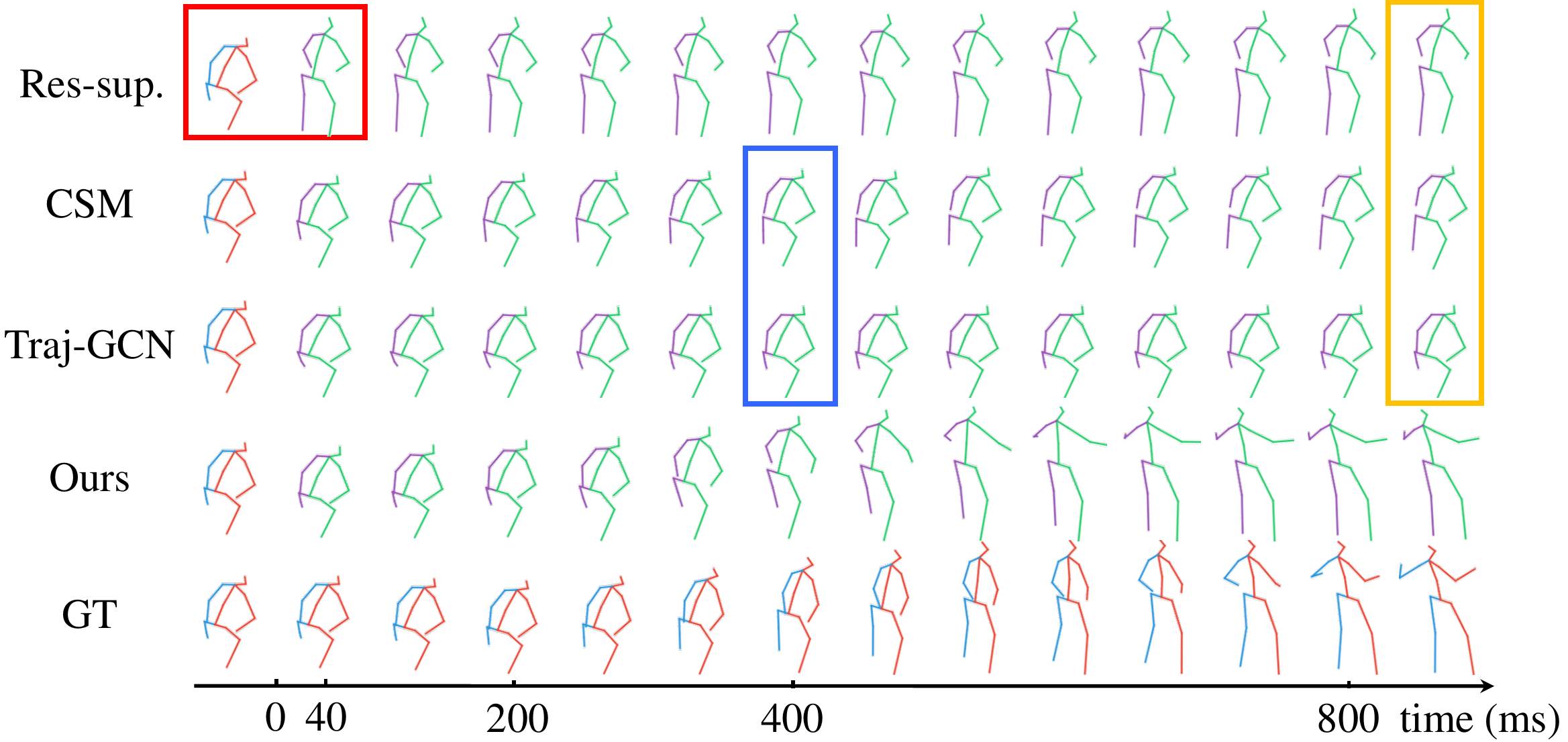}
    \caption{Predicted samples of the action of `Posing' in Human 3.6M dataset from four models in a long term.}
    \label{fig:posing}
\end{figure}

Figure~\ref{fig:posing} illustrates the predicted poses of `Posing' in Human 3.6M in 1000 ms.
We see that the proposed DMGNN could well model the posture, such as stretched bodies and arms; however, Res-sup predicts the motion with large discontinuity between the last observed pose the first predicted one (red box); CSM and Traj-GCN tends to have large errors after the 400th ms (blue box); all the baselines produce unreasonable poses at the 1000th ms (yellow box), which are far from the ground truth.

\begin{figure}[t]
    \centering
    \includegraphics[width=1\columnwidth]{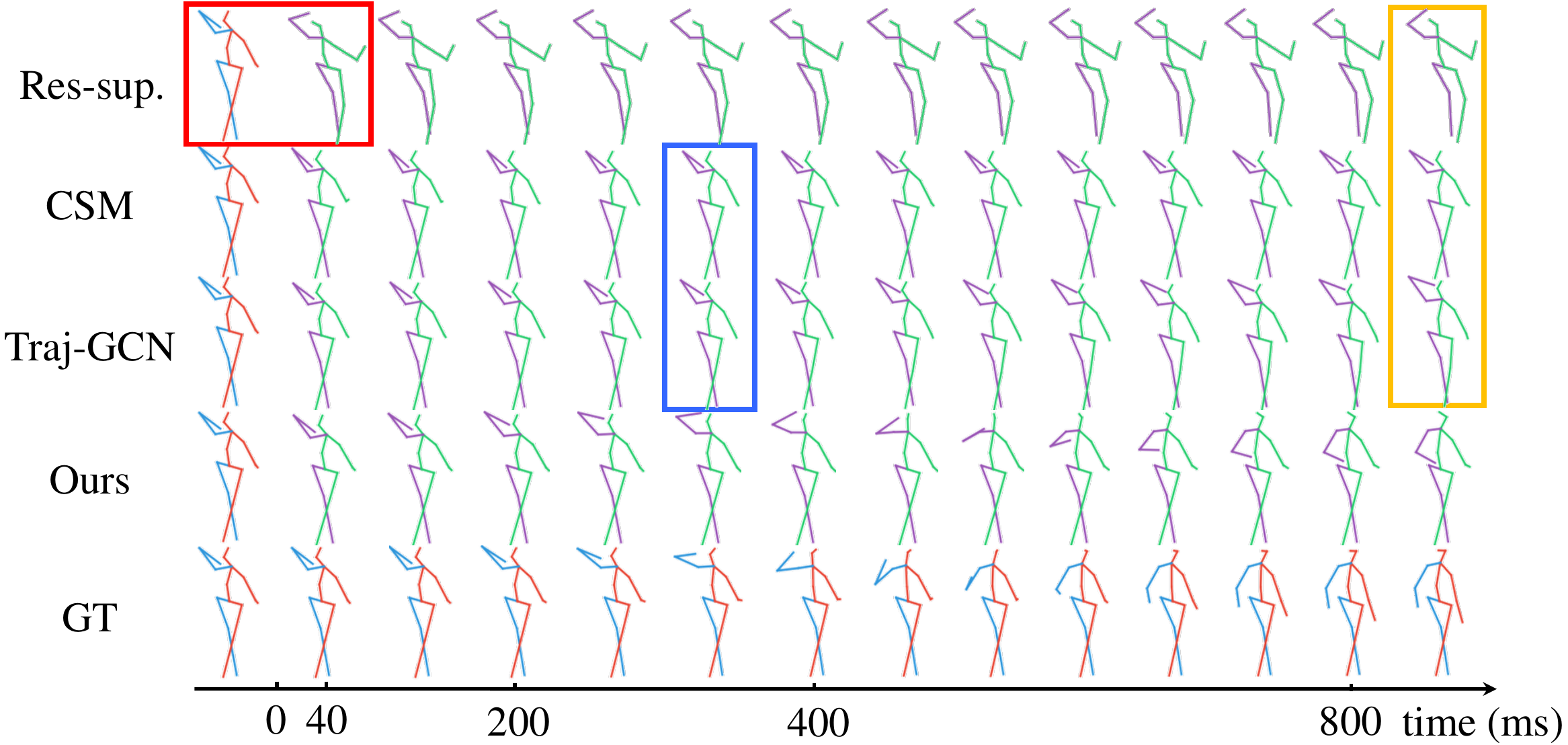}
    \caption{Predicted samples of the action of `Waiting' in Human 3.6M dataset from four models in a long term.}
    \label{fig:waiting}
\end{figure}

We also predict the action of `Waiting' in Human 3.6M in a long term with different methods. The results are illustrated in Figure~\ref{fig:waiting}.
We see that, for baselines, the motion predicted by res-sup has large discontinuity between the last observed pose the first predicted one (red box) and loses the movements, which is far from the ground truths. CSM and Traj-GCN suffer from large errors after the 320th ms; all the baselines predict unreasonable poses at the 1000th ms (yellow box); but the predictions from DMGNN could complete the action reasonably.

\subsection{CMU Mocap Dataset}
We then test DMGNN on the two actions of `Basketball' and `Washing window' in CMU Mocap dataset. The baselines are the CSM~\cite{Li_2018_CVPR} and Traj-GCN~\cite{Mao_2019_ICCV}.

\begin{figure}[t]
    \centering
    \includegraphics[width=1\columnwidth]{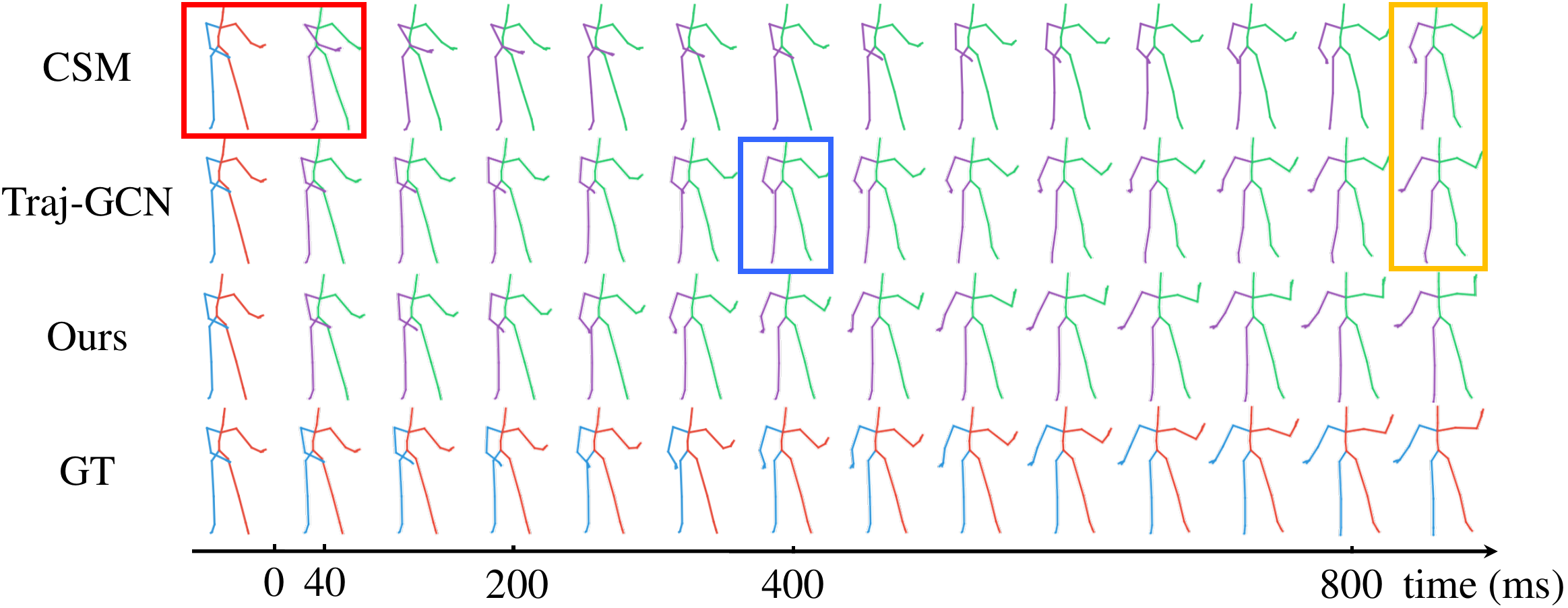}
    \caption{Predicted samples of the action of `Washing window' in CMU Mocap dataset from three models in a long term.}
    \label{fig:washwindow}
\end{figure}

For the action of `Basketball', the main challenge of motion prediction is the running legs and swaying arms. We illustrate the generated samples of three models in Figure~\ref{fig:washwindow}.
\begin{figure}[t]
    \centering
    \includegraphics[width=1\columnwidth]{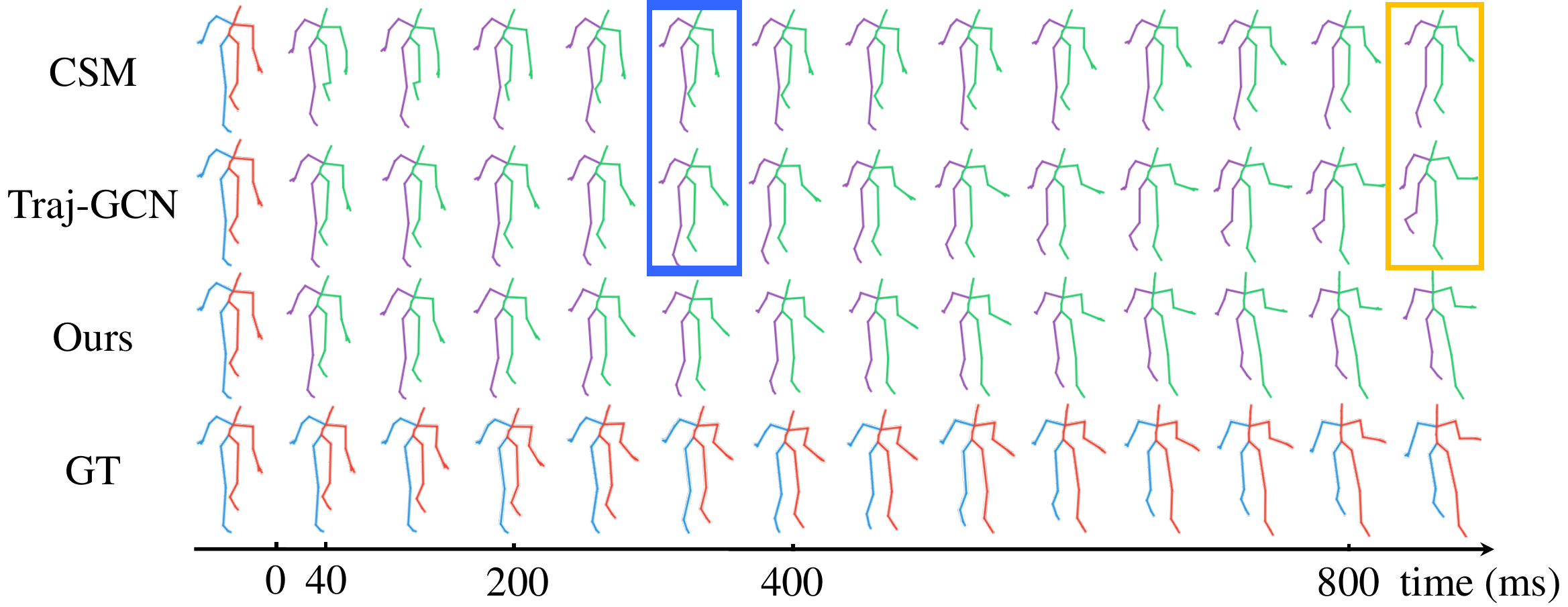}
    \caption{Predicted samples of the action of `Basketball' in CMU Mocap dataset from three models in a long term.}
    \label{fig:basketball}
\end{figure}
We see that the errors of the predictions from CSM and Traj-GCN rise after the 320th ms (blue box); two baselines give unreasonable postures at the 1000th ms in long-term (yellow box); that is, CSM has wrong tilt orientation of the body and the left leg (purple) of the pose predicted by Traj-GCN has inaccurate position; DMGNN could predict motions with smaller errors in both short-term and long-term.

For the action of `Washing window', we also predict the future poses in 1000 ms and illustrate them in Figure~\ref{fig:washwindow}.
We see that the prediction of CSM has large discontinuity between the last observed pose the first predicted one (red box); Traj-GCN tends to have large errors after the 400th ms, since the pose does not raise the left arm (blue box); two baselines give poses at the 1000th ms with large errors (yellow box); but DMGNN could predict motions with smaller errors in both short-term and long-term.
\end{document}